%% file: main.tex
\documentclass[runningheads]{llncs}
\usepackage{eccv}
\usepackage{eccvabbrv}
\usepackage{graphicx}
\usepackage{booktabs}
\usepackage[accsupp]{axessibility} % Improves PDF readability for those with disabilities.
\usepackage{hyperref}
\usepackage{orcidlink}

\usepackage{tabularx}
\usepackage[export]{adjustbox}
\usepackage{varwidth}
\usepackage{array}
\usepackage{makecell}
\usepackage{amsmath}
\usepackage{amssymb}
\usepackage{multirow}
\usepackage{float}
\usepackage{xcolor,subcaption}
\usepackage{placeonpage}
\usepackage{algorithm}
\usepackage{algpseudocode}
\usepackage{cancel}
\begin{document}
% -------------------------------------------------
\title{EBDM: Exemplar-guided Image Translation with Brownian-bridge Diffusion Models}
\titlerunning{EBDM: Exemplar-guided Image Translation with BBDMs.}
\author{
  Eungbean Lee\inst{1}\orcidlink{0000-0003-4839-8540}\and
  Somi Jeong\inst{2}\orcidlink{0000-0002-0906-0988}\and
  Kwanghoon Sohn\inst{1,3}%\orcidlink{0000-0002-3715-0331}
}
\authorrunning{E.~Lee et al.}

\institute{
Yonsei University, Seoul, Korea \\\email{\{eungbean,khsohn\}@yonsei.ac.kr}\and 
NAVER LABS \\\email{somi.jeong@naverlabs.com} \and 
Korea Institute of Science and Technology (KIST), South Korea}
\maketitle
% ---------------------------------------------------------------
% Fig.1 - Motivation
% ---------------------------------------------------------------
\begin{figure*}[bt]
    \centering
    \includegraphics[width=\linewidth]{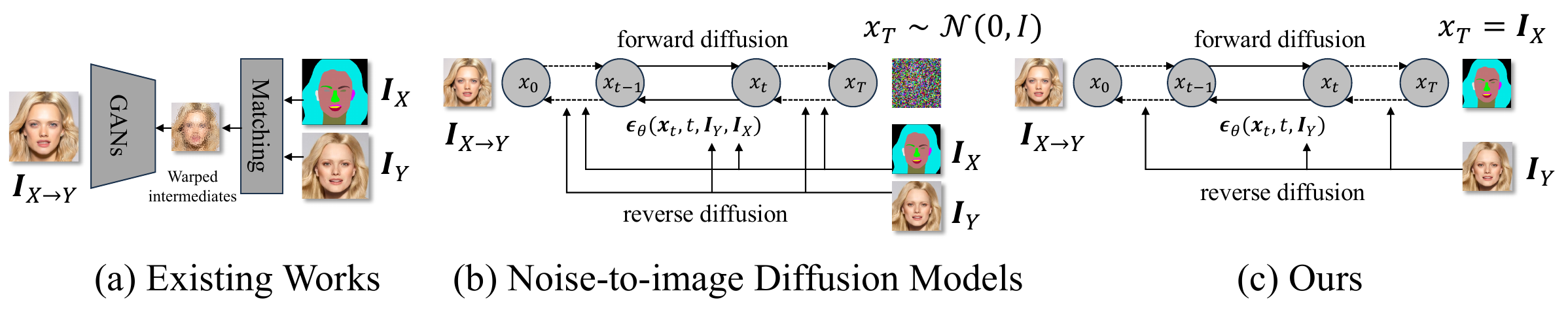}
    \caption{\textbf{Motivation.}(a) Existing methods with matching-than-generation framework, (b) Widely used framework based on conditional noise-to-image diffusion model, and (c) our framework based on Brownian bridge diffusion models.}
    \label{fig:1_motivation}
    \placeonpage{2}
\end{figure*}
% ---------------------------------------------------------------
% Fig.2 - Framework
% ---------------------------------------------------------------
\begin{figure*}[bt]
    \centering
    \includegraphics[width=\linewidth]{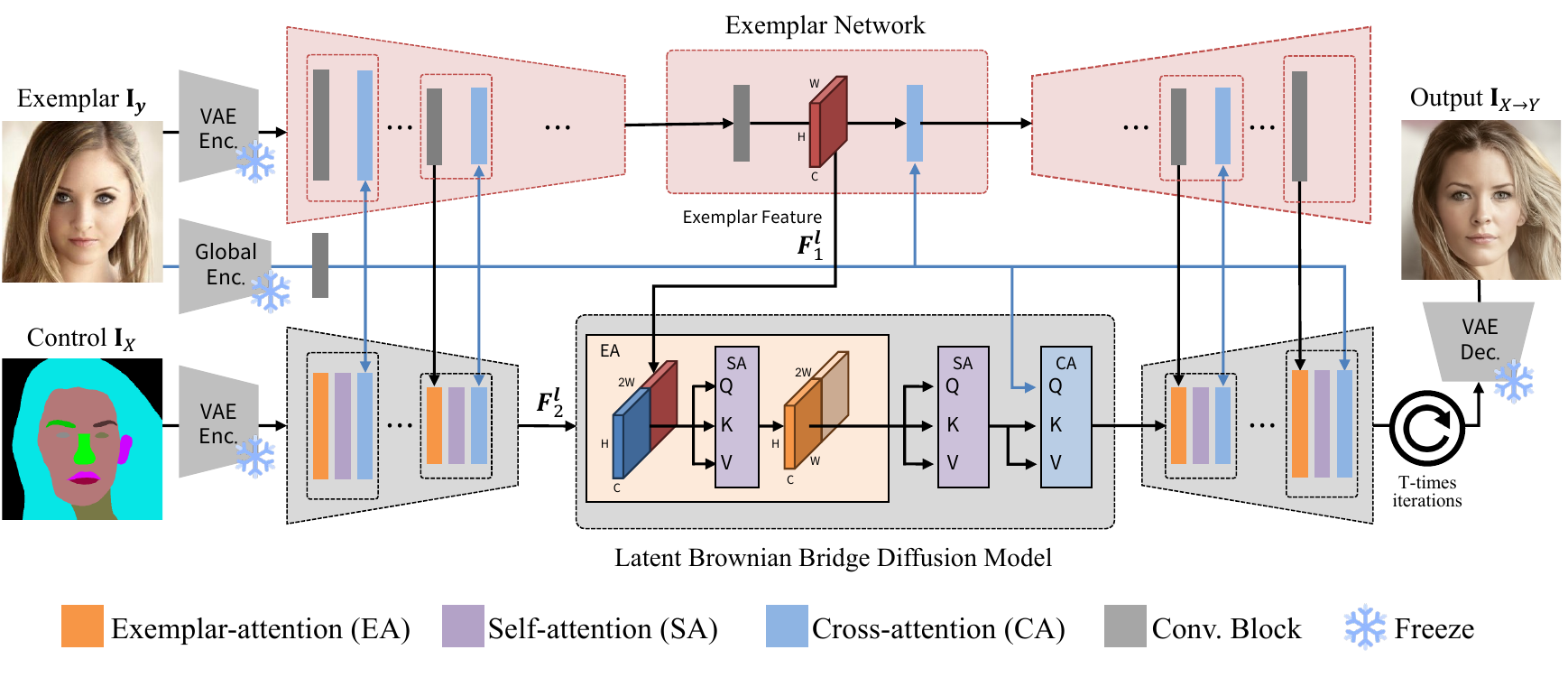}
    \caption{\textbf{Framework overview.} The proposed EBDM framework is a based on (a) Brownian Bridge Diffusion Model and composed of (b) Exemplar Network and a (c) Global Encoder. Global Encoder encodes global style information and Exemplar Network extracts texture information from exemplar image. Extracted texture and global information is then used to guide the diffusion process via Exemplar Attention Module and cross-attention, respectively.}
    \placeonpage{6}
    \label{fig:architecture}
\vspace{-10pt}
\end{figure*}
% ---------------------------------------------------------------
% Fig.3 - Qualitative
% ---------------------------------------------------------------
\begin{figure*}[bt!]
    \centering
    \resizebox{.85\textwidth}{!}{
    \includegraphics{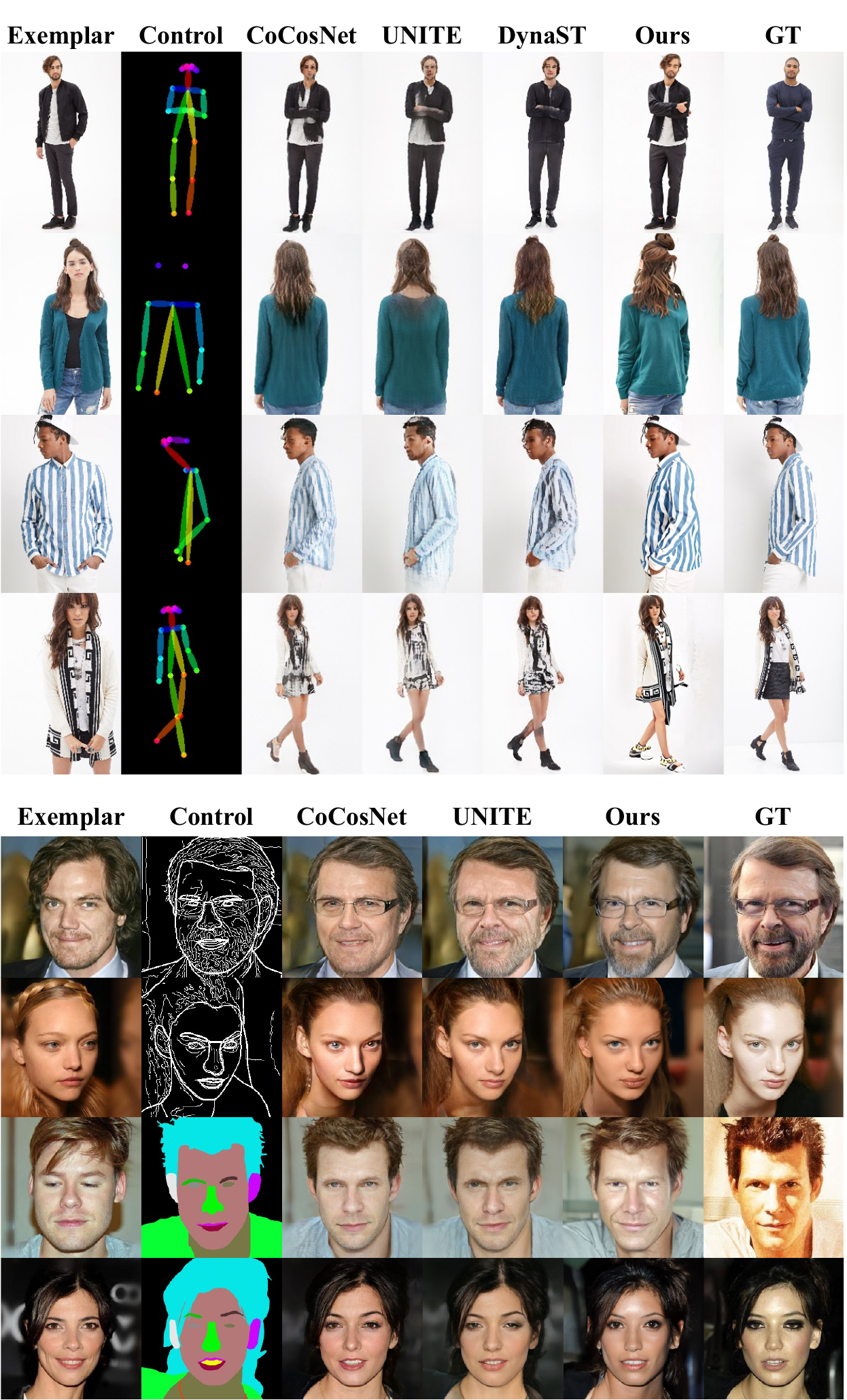}}
    \caption{
        \textbf{Qualitative Results.} Visual comparisons of the proposed EBDM and state-of-the-art methods over three types of exemplar-guided image translation tasks.
        }\label{fig:qualitative}
    \placeonpage{11}
\end{figure*}
% ---------------------------------------------------------------
% Fig.4 - Ablation 1
% ---------------------------------------------------------------
\begin{figure}[!bt]%{width=\linewidth}
    \begin{minipage}[bt!]{0.49\linewidth}
      \centering
      \includegraphics[width=\linewidth]{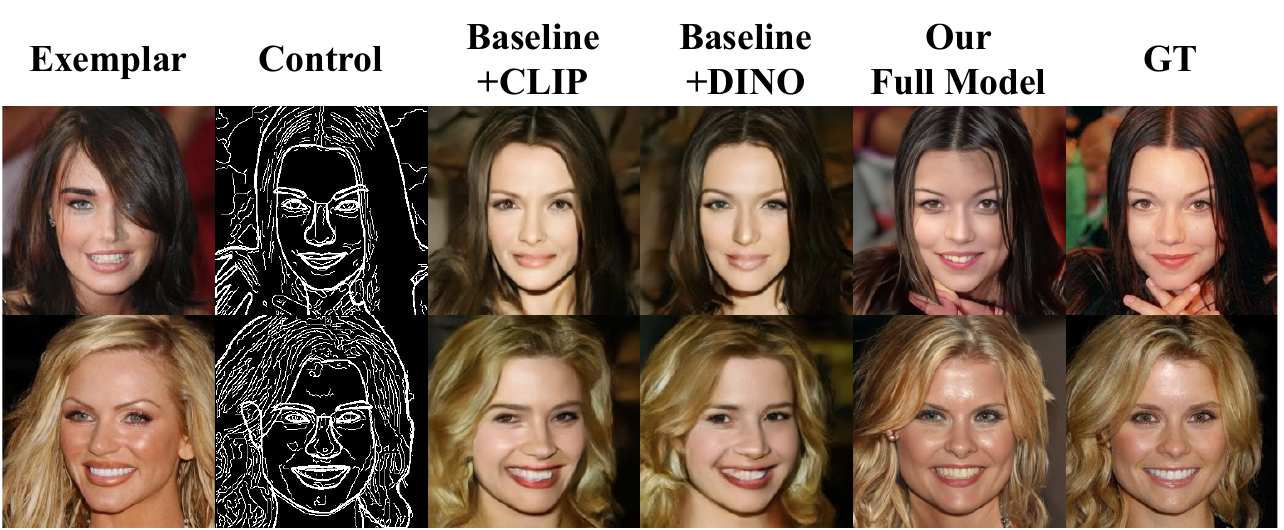}
      \captionof{figure}{Visual Comparison based on the choice of Exemplar Encoder.
      % We compare the impact on selection of Global Encoder, Baseline with CLIP, DINO and our Full Model.
        }\label{fig:vsclip}
    \end{minipage}
    \hfill
    \begin{minipage}[bt]{0.49\linewidth}
    \resizebox{.9\linewidth}{!}{
      \centering
      \begin{tabular}{c ccc}
      \\
      \hline
        \textbf{Method} & SSIM$\uparrow$&FID$\downarrow$&Sem.$\uparrow$ \\ \hline
        \textbf{Baseline            } & 0.831 & 16.31 & 0.531\\
        \textbf{Baseline+CLIP       } & 0.632 & 23.42 & 0.752\\
        \textbf{Baseline+DINO       } & 0.754 & 21.32 & 0.786\\
        \hline
        \textbf{Ours}            & \textbf{0.901} & \textbf{11.84}& \textbf{0.920} \\
        \hline
        \end{tabular}
        }
        \captionof{table}{Quantitative Results from the Ablation Study.} \label{tab:vsclip}
      \end{minipage}
      \placeonpage{13}
% \vspace{-10pt}
\end{figure}
% ---------------------------------------------------------------
% Fig.5 Ablation 2
% ---------------------------------------------------------------
\begin{figure}[!bt]%{width=\linewidth}
    \begin{minipage}[bt!]{0.49\linewidth}
      \centering
      \includegraphics[width=\linewidth]{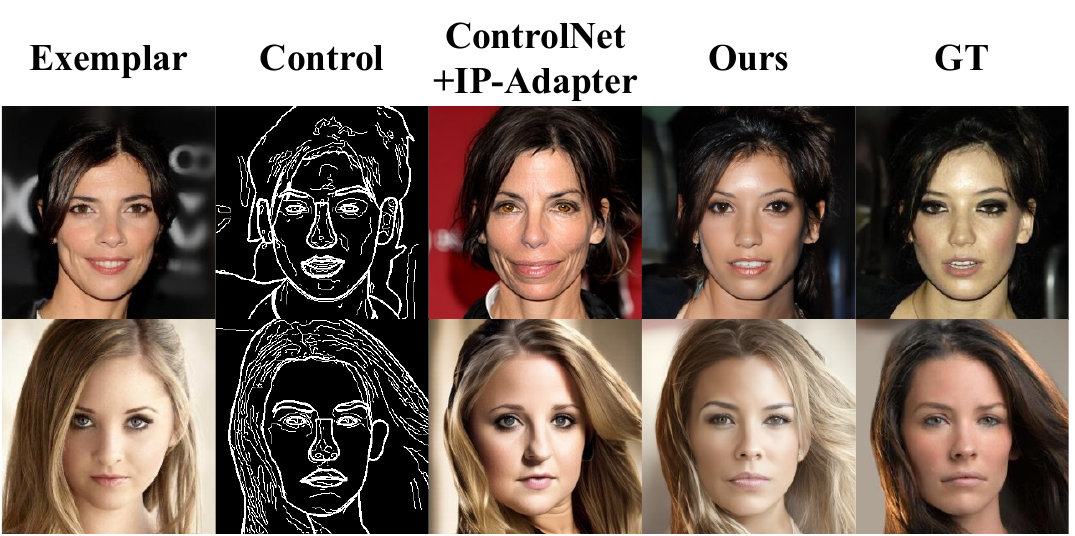}
      \captionof{figure}{
        Qualitative Comparison to SOTA Diffusion-based Model. % We compare our method to state-of-the-art method combining various tequniques, Stable Diffusion, ControlNet and IP-Adapter.
      }\label{fig:ablation-vs-diffusion}
    \end{minipage}
    \hfill
    \begin{minipage}[bt]{0.49\linewidth}
    \resizebox{.9\linewidth}{!}{
      \centering
      \begin{tabular}{c cc}
      \\ % blank line to align with figure.
      \hline
        % \multirow{2}{*}{\textbf{Method}} &
        % \multicolumn{2}{c}{\textbf{CelebA-HQ (Edge)}} \\
        % \cline { 2 - 3 }
        % \multicolumn{3}{c}{\textbf{CelebA-HQ (Edge)}}\\ \hline
        \textbf{Method} & SSIM$\uparrow$&PSNR$\downarrow$\\ \hline
        \textbf{ControlNet     } & 0.882 & 35.30 \\
        \textbf{ControlNet+CLIP} & 0.894 & 35.94 \\
        \hline
        \textbf{Ours}            & \textbf{0.901} & \textbf{36.40} \\
        \hline
        \end{tabular}
    }
        \captionof{table}{Quantitative comparison to SOTA diffusion-based model.} \label{tab:ablation-vs-diffusion}
      \end{minipage}
      \placeonpage{13}
\vspace{-10pt}
\end{figure}
% ---------------------------------------------------------------
% Abstract 
% ---------------------------------------------------------------
\begin{abstract}
Exemplar-guided image translation, synthesizing photo-realistic images that conform to both structural control and style exemplars, is attracting attention due to its ability to enhance user control over style manipulation. 
Previous methodologies have predominantly depended on establishing dense correspondences across cross-domain inputs. 
Despite these efforts, they incur quadratic memory and computational costs for establishing dense correspondence, resulting in limited versatility and performance degradation.
In this paper, we propose a novel approach termed Exemplar-guided Image Translation with Brownian-Bridge Diffusion Models (EBDM). 
Our method formulates the task as a stochastic Brownian bridge process, 
a diffusion process with a fixed initial point as structure control and translates into the corresponding photo-realistic image while being conditioned solely on the given exemplar image. 
To efficiently guide the diffusion process toward the style of exemplar, 
we delineate three pivotal components: the Global Encoder, the Exemplar Network, and the Exemplar Attention Module to incorporate global and detailed texture information from exemplar images.
Leveraging Bridge diffusion, the network can translate images from structure control while exclusively conditioned on the exemplar style, leading to more robust training and inference processes. 
We illustrate the superiority of our method over competing approaches through comprehensive benchmark evaluations and visual results. 
% \keywords{Generative model\and Image Synthesis\and Diffusion Models}
\keywords{Generative model\and Image Translation\and Diffusion Models \and Image Synthesis} 
\end{abstract}
%
%
%
%
% ---------------------------------------------------------------
% Intro 
% ---------------------------------------------------------------
\section{Introduction}\label{sec:intro}
The rising interest in applications of image synthesis has led to a notable surge in demand for image generation capabilities that extend beyond text prompts, emphasizing control through exemplar images or structured inputs. Exemplar-guided image translation task aims to generate photo-realistic images conditioned on both a style exemplar image and specific structural controls, such as segmentation masks, edge maps, or pose keypoints.

To synthesize images guided by the style of an exemplar and structure controls, pioneer works~\cite{pix2pix, pix2pixhd, spade, clade, inade} have emerged. 
Formulated as an ill-posed problem, these methods globally leverage the style of an exemplar. Despite promising results, these methods overlook local details, which leads to compromised generation quality.

To enhance the capture of local styles from exemplar, significant efforts~\cite{cocosnet, cocosnetv2, unite, rabit, dynast, mclnet, midms} have been explored to establish cross-domain correspondences between input control and exemplar images, thereby imposing local style through a matching process.
Zhang \etal.~\cite{cocosnet} have explored the construction of cross-domain correspondences using cosine similarity for exemplar-based image translation and subsequent works~\cite{cocosnetv2, unite, rabit, mclnet} introduced various techniques to reduce computational complexity and address many-to-one matching problems.
Moreover, they predominantly capture style information at a coarse scale, which leads to performance degradation, because their methods are significantly influenced by the quality of warped intermediate features derived from sparse correspondences between two domains, often failing to accurately reflect the dynamic nature of matching.
This failure leads to local distortion, blurred details, and semantic inconsistency. Furthermore, models leveraging Generative Adversarial Networks (GANs) face the intrinsic limitations of GANs, such as mode collapse, limited diversity, and the out-of-range problem~\cite{bdinvert, styleheat}.

Recently, diffusion models~\cite{ddpm, sdm, ddim, ldm}, which generate high-quality images through iterative denoising processes, have attained significant success in the field of image synthesis for their several advantages, including broader distribution coverage, more stable training, and enhanced scalability compared to GANs~\cite{ddpm, guideddiffusion}.
As the surge in demand for customized image generation has advanced, Text-to-image (T2I) synthesis, conditioned on text prompts, has been extensively explored in works such as~\cite{glide, dalle2, vqdiffusion}.
Beyond mere text prompts, numerous studies have sought solutions to address precise style through model fine-tuning~\cite{dreambooth, lora}, prompt engineering~\cite{promptengineering, zeroshotprompt}.
Additionally, efforts have been made to incorporate structure controls (\ie, edge, depth, mask, pose, \textit{etc}.)~\cite{controlnet, unicontrolnet, t2iadapter, composer, piti} as generative guidance. 

Although diffusion models have demonstrated impressive performance, exemplar-based image translation remains largely unexplored.
First, it is challenging to find an accurate prompt that conveys every desired aspect of an image.
Second, it is hard to address exemplar style because fine-tuning offers quality at a high cost, prompt engineering is more affordable but less detailed, and CLIP representations are not sufficient to address all details in visual cues.
Lastly, achieving simultaneous conditioning on both style exemplars and structured controls is challenging, particularly because the diffusion process used in such tasks can be highly sensitive to hyperparameters, including the guidance scales of structure control and embeddings.

To solve the above issues, we introduce a sophisticated technique called EBDMs (Exemplar-based image translation with Brownian bridge Diffusion Models) that fully leverages diffusion models.
Our method leverages a stochastic Brownian bridge process~\cite{bbdm} that directly learns translation between two domains, thus generating images from structure controls without any conditioning mechanism.
To explore desired style control, we propose a Global Encoder and Exemplar Network to leverage coarse and fine details from exemplar images.
Moreover, the Exemplar Attention Module effectively consilates the texture information from the exemplar into the denoising process.
Our method can generate images by conditioning on structure control and style exemplars with single conditioning simultaneously.
We conduct extensive experiments on various datasets, including mask-to-image, edge-to-image~\cite{celebahq}, and keypoint-to-image~\cite{deepfashion}.
The experimental results demonstrate the superiority of our approach not only performance but also computational efficiency.
The contribution of this work can be summarized as follows:
\begin{itemize}
    \item We introduce the EBDMs, a novel framework leveraging the stochastic Brownian Bridge diffusion process that translates from structure control to a photo-realistic image while effectively exploiting style from exemplars.
    \item The proposed method formulated the problem into a single-conditioned bridge diffusion process that ensures the training and inference more robust.
    \item We propose Global Encoder, Exemplar Network, and Exemplar Attention Module to address both global style and detailed texture of exemplar image.
    \item Extensive experiments demonstrate that our approach achieves favorable performance on various exemplar-guided image translation tasks.
\end{itemize}
%
%
%
% ---------------------------------------------------------------
% Related Works
% ---------------------------------------------------------------
\section{Related Works}\label{sec:related_works}
\subsection{Controllable Diffusion Models}
% a) Diffusion models
Diffusion models~\cite{ddpm, sdm, ddim, guideddiffusion, ldm} aim to synthesize images from random Gaussian noise via an iterative denoising process. 
% b) T2I
For customized image generation, recent methods have explored text-guided image generation (T2I)~\cite{glide, vqdiffusion, dalle2, ldm, imagen} and demonstrated extraordinary generative capabilities in modeling the intricacies of complex images.
GLIDE~\cite{glide} aggregated the CLIP texture representations utilizing classifier-free guidance~\cite{cfg}. 
DALLE-2~\cite{dalle2} proposed a cascade model using the CLIP latent.
VQ-Diffusion~\cite{vqdiffusion} proposed to learn the diffusion process on the discrete latent space of VQ-VAE~\cite{vqvae}.

% c) Structure control - controlnet, unicontrol
To address the structure controls (such as mask, edge, pose, \textit{etc}.), several works have proposed fine-tuning approach~\cite{piti} or adaptive models~\cite{controlnet, unicontrolnet, t2iadapter} in addition to text prompts.
ControlNet~\cite{controlnet} proposed an adaptive network to provide structure guidance to T2I models followed by Uni-ControlNet~\cite{unicontrolnet} which expands to a unified framework to accept diverse control signals at once.
Concurrently, T2I-Adapter~\cite{t2iadapter} introduces a more simple and lightweight adapter.
Such methods enable to provide the structural guidance to existing T2I diffusion models thus providing more precise spatial control.

% d) Exemplar style - finetuning, prompt engineering
On the other hand, to accurately reflect the style of an exemplar, 
numerous studies have been conducted such as model fine-tuning~\cite{dreambooth, imagic, lora, ipadapter, textualinversion}, prompt engineering~\cite{promptengineering, zeroshotprompt, promptguidelines}.
DreamBooth~\cite{dreambooth} proposed to fine-tune the T2I models with exemplar image and LoRA~\cite{lora} proposed a more effective tuning method.
IP-Adapter~\cite{ipadapter} proposed decoupled cross-attention to effectively inject exemplar image features into the denoising network.
Moreover, Guo \etal~\cite{zeroshotprompt} proposed the image-specific prompt learning method to learn domain-specific prompt vectors.
While other methods~\cite{ptp, sdedit, nti, palette} enable zero-shot editing of an exemplar image based on a target caption.
Despite these method capabilities, it is challenging to find the prompt that accurately generates the image a user envisions, mainly because effectively reflects all desired aspects of an image through text, especially those that are difficult or impossible to describe precisely.

\subsection{Exemplar-guided Image Translation}
The exemplar-guided image translation task involves generating an image based on an input exemplar and structure controls such as an edge, pose, or mask.
A major challenge lies in effectively guiding the context within exemplars relative to the input controls.
% a) Global manners 
The SPADE~\cite{spade} framework proposed spatially-adaptive normalization to generate an image from the semantic mask followed by class-adaptive~\cite{clade} and instance-adaptive~\cite{inade}.
While these approaches have shown promising in global-style translation, they overlooked local details compromising generation quality.

% b) Local manner: exemplar-guided image translation
To address the local details, significant efforts have been focused on building dense correspondence.
Zhang~\etal~\cite{cocosnet} proposed building dense correspondence between input semantic and exemplar image.
Although their method has shown promising results, their method is limited by many-to-one matching issues and the quadratic computational and memory complexities of dense matching operations, restricting it to capturing only coarse-scale warped features.
To alleviate these issues, recent works introduced effective correspondence learning such as 
GRU-assisted Patch-Match~\cite{cocosnetv2}, 
unbalanced optimal transport~\cite{unite}, 
bi-level feature alignment strategy~\cite{rabit}, 
multi-scale dynamic sparse attention~\cite{dynast},
Cross-domain Feature Fusion Transformer~\cite{cfftgan} 
and Masked Adaptive Transformer~\cite{matebit}.
Although they have demonstrated promising results, their matching-based framework still suffers from inherent problems such as sparse matching.

% c) diffusion models
Meanwhile, recent progress~\cite{midms, composer, paint-by-example,imagebrush} has leveraged diffusion models to bridge the gap between style exemplars and structural controls.
Seo~\etal~\cite{midms} proposed a two-staged framework, which is a matching module followed by a diffusion module.
Although they have successfully applied diffusion models, they still heavily rely on a matching-based framework that does not fully utilize the diffusion models.
Paint-by-example~\cite{paint-by-example} proposed self-supervised training for image disentanglement and reorganization, while Composer~\cite{composer} conceptualized an image as a composition of several representations, suggesting a decompose-then-recompose approach
and ImageBrush~\cite{imagebrush} learns visual instructions.
Although they have demonstrated promising results, they offer limited control, typically constrained to structure-preserving appearance changes or uncontrolled image-to-image translation.
%
% ---------------------------------------------------------------
% Prelinamaries
% ---------------------------------------------------------------
\section{Preliminaries}\label{preliminaries}
\subsection{Diffusion Models}
The general idea of Denoising Diffusion Probabilistic Model (DDPM)~\cite{ddpm} is to generate images from Gaussian noise via $T$ steps of an iterative denoising process.
It consists of two processes: the forward process and the reverse process.
Given the original data $\boldsymbol{x}_0 \sim q_{data}\left(\boldsymbol{x}_0\right)$, the forward diffusion process maps $\boldsymbol{x}_0$ into noisy latent variables $\{\boldsymbol{x}_t\}_{t=0}^T$ 
can be obtained as: 
$\boldsymbol{x}_t=\sqrt{\alpha_t} \boldsymbol{x}_0+\sqrt{1-\alpha_t} \boldsymbol{\epsilon}$
where $\boldsymbol{\epsilon}$ is the Gaussian noise and $\{\alpha_t\}^T_{t=0}$ is pre-defined schedule.
% b) Reverse Process
On the other hand, the corresponding reverse process aims to predict the original data $\boldsymbol{x}_0$ starting from the pure Gaussian noise $\boldsymbol{x}_T \sim \mathcal{N}(\mathbf{0}, \textbf{I})$ through iterative denoising processes with pre-defined time steps.
It is formulated as another Markov chain as $p_\theta\left(\boldsymbol{x}_{t-1} \mid \boldsymbol{x}_t\right):=\mathcal{N}\left(\boldsymbol{x}_{t-1} ; \boldsymbol{\mu}_\theta\left(\boldsymbol{x}_t, t\right), \sigma_t^2 \textbf{I}\right)$ with learned mean and fixed variance.
% c) training objectives
The denoising network $\boldsymbol{\epsilon}_\theta$ is trained to predict the noise by minimizing a weighted mean squared error loss, defined as:
\begin{equation}
\label{eq:elbo-ddpm}
L(\theta)=\mathbb{E}_{t,\boldsymbol{x}_0, \boldsymbol{\epsilon}}[\left\|\boldsymbol{\epsilon}-\boldsymbol{\epsilon}_\theta\left(\boldsymbol{x}_t, t\right)\right\|_2^2].
\end{equation}
% d) conditional
Similarly, the conditional diffusion models~\cite{palette, unit-ddpm} directly inject the condition $\boldsymbol{y}$ into the training objective (eq.~\ref{eq:elbo-ddpm}), such as $L(\theta)=\mathbb{E}_{t, \boldsymbol{x}_0, \boldsymbol{\epsilon}}\left\|\boldsymbol{\epsilon}-\boldsymbol{\epsilon}_\theta\left(\boldsymbol{x}_t, \boldsymbol{y}, t\right)\right\|_2^2$.

\subsection{Brownian Bridge Diffusion Models}
% a) brownian bridge
A Brownian Bridge Diffusion Model (BBDM)~\cite{bbdm} is an image-to-image translation framework based on a stochastic Brownian Bridge process.
Unlike DDPM that conclude at Gaussian noise $\boldsymbol{x}_T\sim \mathcal{N}(0, \textbf{I})$, 
BBDM assumes that both endpoints of the diffusion process as fixed data points from an arbitrary joint distribution, \ie $(\boldsymbol{x}_T,\boldsymbol{x}_0)\sim q_\text{data}(\mathcal{X},\mathcal{Y})$.
The BBDM directly learns image-to-image translation $q(\boldsymbol{x}_0|\boldsymbol{x}_T)$ with boundary distribution $q_\text{data}(\boldsymbol{x}_0,\boldsymbol{x}_T)$ independent of any conditional process, that enhances the fidelity and diversity of the generated samples.
% b) forward process
The forward process of the Brownian Bridge forms a bridge between two fixed endpoints at $t=0$ and $T$:
\begin{equation}\label{bbdm-forward}
q\left(\boldsymbol{x}_t \mid \boldsymbol{x}_0, \boldsymbol{y}\right)=\mathcal{N}\left(\boldsymbol{x}_t ;\left(1-m_t\right)\boldsymbol{x}_0+m_t \boldsymbol{y}, \delta_t \textbf{I}\right),
\quad \text{where}\quad \boldsymbol{y}=\boldsymbol{x}_T
\end{equation}
where $m_t=t/T$ and variance term $\delta_t=2(m_t-m_t^2)$.

% b) Reverse Process
The reverse process of BBDM aims to predict $\boldsymbol{x}_{t-1}$ given $\boldsymbol{x}_t$: 
\begin{equation}\label{eq:mean-bbdm}
p_\theta(\boldsymbol{x}_{t-1} \mid \boldsymbol{x}_t, \boldsymbol{y})
=\mathcal{N}(\boldsymbol{x}_{t-1} ; \boldsymbol{\mu}_\theta\left(\boldsymbol{x}_t, t\right), \tilde{\delta}_t \textbf{I}),
\end{equation}
where $\tilde{\delta}_t$ is the variance of Gaussian noise at step $t$ and $\boldsymbol{\mu}_\theta\left(\boldsymbol{x}_t, t\right)$ is the predicted mean value of the noise, which is network to be learned.
% c) Training Objective
The training objective of BBDM is optimizing the Evidence Lower Bound (ELBO), simplified as:
\begin{equation}\label{eq:elbo-bbdm}
\mathbb{E}_{\boldsymbol{x}_0, \boldsymbol{y}, \boldsymbol{\epsilon}}
\left[
c_{\epsilon t} %$ <<<<<<< we earase it b/c c_{\epsilon t} = 1/t 
\|m_t(\boldsymbol{y}-\boldsymbol{x}_0\right)+\sqrt{\delta_t} \boldsymbol{\epsilon}-\boldsymbol{\epsilon}_\theta(\boldsymbol{x}_t, t)\|^2].
\end{equation}
where $c_{\epsilon t}$ is the coefficient term of estimated noise $\boldsymbol{\epsilon_\theta}$ in mean value term, $\tilde{\boldsymbol{\mu}}_t$.
%
%
%
% ---------------------------------------------------------------
% Proposed Method 
% ---------------------------------------------------------------
\section{Methodology}\label{methodology}
% ========================================================================
In this section, we delineate our framework based upon discrete-time stochastic Brownian Bridge diffusion process~\cite{bbdm} for Exemplar-guided image translation (~\cref{fig:architecture}).
Given a control $\boldsymbol{I}_\mathcal{X}$ sampled from domain $\mathcal{X}$ alongside an exemplar image $\boldsymbol{I}_\mathcal{Y}$ from domain $\mathcal{Y}$,
the primary objective is to generate a target image $\boldsymbol{I}_{\mathcal{X}\rightarrow \mathcal{Y}}$
that retains the structure of $\boldsymbol{I}_\mathcal{X}$ embodying the style of $\boldsymbol{I}_\mathcal{Y}$.
The key to our method is the infusion of style information from $\boldsymbol{I}_\mathcal{Y}$ to guide the diffusion trajectory of the target image.
To facilitate this, our method integrates three components: a denoising network equipped with an Exemplar Attention Module, a Global Encoder, and Exemplar Network.
The Global Encoder extracts global style information of $\boldsymbol{I}_\mathcal{Y}$ and 
the Exemplar Network captures the appearance features of  $\boldsymbol{I}_\mathcal{Y}$.  
The Exemplar Attention Module selectively incorporates appearance information into the denoising process.

In the following sections, 
we present a detailed explanation of the framework (Sec.~\ref{sec:method}), training strategy (Sec.~\ref{sec:training}), and sampling strategy (Sec.~\ref{sec:sampling}).

\subsection{Exemplar-guided Brownian Bridge Diffusion Models}\label{sec:method}
\noindent\textbf{Denoising Network. }
Employing the Brownian Bridge diffusion process, our denoising U-Net directly learns the translation from the input controls  $\boldsymbol{I}_\mathcal{X}$ to images $\boldsymbol{I}_{\mathcal{X}\rightarrow \mathcal{Y}}$ that preserves the structures of controls.
For efficient training and inference, we employ the Stable Diffusion~\cite{ldm} framework.
Specifically, given an image $\boldsymbol{I}$, the encoder $\mathcal{E}$ maps it into a latent space $\boldsymbol{z}=\mathcal{E}\left( \boldsymbol{I}\right)$, subsequently reconstructed by the decoder $\hat{\boldsymbol{I}}=\mathcal{D}\left( \boldsymbol{z}\right)$.
The denoising U-Net $\boldsymbol{\epsilon}_\theta$ learns to establish the bridge from fixed initial point $\boldsymbol{x}_T=\boldsymbol{z}_\mathcal{X}$ to the target,  $\boldsymbol{x}_0=\boldsymbol{z}_{\mathcal{X}\rightarrow \mathcal{Y}}$.

Unlike existing noise-to-image diffusion frameworks~\cite{controlnet,t2iadapter} embed the structural information through intricate frameworks, 
our approach translates from structural control to images without explicit conditional operation.
Consequently, our framework is able to solely focus on exemplar information that fosters enhanced stable training and inference performance.

\noindent\textbf{Global Encoder. } 
The Global Encoder, utilizing \texttt{DINOv2}~\cite{dinov2}, captures the global style information from the exemplar image $\boldsymbol{I}_\mathcal{Y}$.
Specifically, exemplar image $\boldsymbol{I}_\mathcal{Y}$ is processed through the Global Encoder, subsequently, $\texttt{[CLS]}$ token is extracted and passed through a linear layer to encapsulate global style attributes:
\begin{equation}
    \tau_\theta(\boldsymbol{I}_\mathcal{Y}) = \texttt{Linear}\left(\texttt{DINO}(\boldsymbol{I}_\mathcal{Y})_\texttt{[CLS]}\right) \in \mathbb{R}^{c},
\end{equation}
where $c$ denotes the dimension of $\texttt{[CLS]}$ token.
The global features are utilized as global style information through a cross-attention mechanism, ensuring that the synthesized output accurately reflects the exemplar's global style.

In the context of text-to-image synthesis~\cite{ddpm,ldm}, 
prior works have extensively leveraged the \texttt{CLIP} image encoder to convey high-level semantic prompts via cross-attention.
This approach, however, primarily focuses on the semantic alignment of prompts and images, thereby overlooking the representation of detailed textures.
Furthermore, our method does not need textual prompt alignment.
Motivated by recent studies~\cite{splicingvit, diffuseit} that have demonstrated the superior proficiency of \texttt{DINO}~\cite{dino, dinov2} over \texttt{CLIP}~\cite{clip} in encapsulating a broader capability of semantic features in images, attributed to its self-supervised learning strategy,
our method incorporates the use of a pre-trained \texttt{DINOv2} encoder to enhance the semantic fidelity of generated images.

\noindent\textbf{Exemplar Network. }
Notwithstanding the capability of Global Encoder in capturing overarching style information, 
it is limited to the retention of fine-grained details because it encodes exemplar in low resolution ($224^2$).
In contrast, the exemplar-guided image translation tasks require higher fidelity to detail.
To this end, we introduce Exemplar Network, referred to  $\psi_\theta$, of which the objective is to capture the detailed texture information from the exemplar image, thereby compensating for the global information.

The Exemplar Network adopts a siamese configuration akin to a denoising U-Net, 
streamlined by omitting the redundant layers for enhanced efficiency during training and inference.
It encodes the exemplar $\boldsymbol{z}_\mathcal{Y}$ into a feature maps $\{\boldsymbol{F}^l_1\}^N_{l=0}$ across $N$ blocks.
Additionally, it processes the global information through cross-attention mechanisms in each block.
The exemplar features $\{\boldsymbol{F}^l_1\}^N_{l=0}$ are then integrated into the noise prediction branch via Exemplar Attention Module.
 
\noindent\textbf{Exemplar Attention module.}
The straightforward approaches to integrate additional features into the denoising network are concatenation~\cite{controlnet, unicontrolnet} or addition~\cite{t2iadapter} % as existing works proposed.
However, in contrast to existing works in that control features are spatially aligned with the target image, 
this approach is not suitable for our task because the exemplar image and target control are not spatially aligned.
Therefore, we propose an Exemplar Attention Module to integrate the exemplar features from the Exemplar Network, $\boldsymbol{F}_1^l \in \mathbb{R}^{C\times H\times W}$, into noise prediction features, $\boldsymbol{F}_2^l \in \mathbb{R}^{C\times H\times W}$ for each $l$ block.
First, these features are concatenated into a spatial-wise: 
$\boldsymbol{F}^l_{\text{in}}=\texttt{concat}(\boldsymbol{F}^l_1,\boldsymbol{F}^l_2)\in \mathbb{R}^{C\times H \times 2W}$.
Following this, self-attention is applied to compute the spatial attention across the features:
\begin{equation}
\begin{aligned}
\boldsymbol{Q}=\phi^l_q(\boldsymbol{F}_{in}^l),\quad
\boldsymbol{K}=\phi^l_k(\boldsymbol{F}_{in}^l),\quad
\boldsymbol{V}=\phi^l_v(\boldsymbol{F}_{in}^l) \\
\boldsymbol{F}^l_{\text{att}}=\frac{\boldsymbol{Q}\boldsymbol{K}^T}{\sqrt{\boldsymbol{V}}},\quad 
\boldsymbol{F}^l_{\text{EA}}=\boldsymbol{W}^l\text{Softmax}(\boldsymbol{F}^l_{att})\boldsymbol{V}+\boldsymbol{F}^l_{\text{in}}
\end{aligned}
\end{equation}
where $\boldsymbol{Q}, \boldsymbol{K}$ and $\boldsymbol{V}$ represents query, key and value, respectively, $\phi\left( \cdot \right)$ is layer-specific $1\times1$ convolution operation, and $\boldsymbol{W}^l$ is trainable parameter.
Subsequently, exemplar-attended feature $\boldsymbol{F}^l_{\text{EA}} \in \mathbb{R}^{C\times H\times 2W}$ is segmented, with portions corresponding to the 
 denoising features are extracted and forwarded toward the output, 
$\boldsymbol{F}^{l}_{\text{out}} = \text{Chunk}(\boldsymbol{F}^l_{\text{EA}}, 2, \text{dim=0}) \in \mathbb{R}^{C\times H\times W}$.
The Exemplar Attention Module computes the region of interest for each query position, a crucial step in effectively directing the denoising steps towards the target exemplar style. 
This approach enables the denoising process to selectively assimilate features from the Exemplar Network, enhancing the fidelity of the output to the desired stylistic attributes.

% ==========================================================================================
% ==========================================================================================
\noindent\textbf{Training Objectives. }\label{sec:objective}
The training process is performed by optimizing the Evidence Lower Bound (ELBO), following BBDM~\cite{bbdm}, where the marginal distribution is conditioned on $\boldsymbol{x}_\mathcal{T}$.
Thus, the training objective ELBO in eq.~\ref{eq:elbo-bbdm} can be simplified as:
\begin{equation}
    \mathbb{E}_{\boldsymbol{x}_0, \boldsymbol{y}, \boldsymbol{I}_\mathcal{Y}, \boldsymbol{\epsilon}}\left[
    c_{\epsilon t}
    \left\|m_t\left(\boldsymbol{x}_T-\boldsymbol{x}_0\right)+\sqrt{\delta_t} \boldsymbol{\epsilon}-\boldsymbol{\epsilon}_\theta\left(\boldsymbol{x}_t, t,\tau_\theta(\boldsymbol{I}_\mathcal{Y}),\psi_\theta(\boldsymbol{z}_\mathcal{Y}, \tau_\theta(\boldsymbol{I}_\mathcal{Y})\right)\right\|^2\right],
\end{equation}
where $c_{\epsilon t}$ is the loss weighting function that develops into $1/t$, and $\delta_t$ denotes the preserved variance schedule, $\delta_t=2(m_t -m_t^2)$.

\input{tab/algorithm}
\subsection{Training Strategy}\label{sec:training}
The training process is unfolded in two stages.
In the first stage, denoising U-Net, which utilizes the Global Encoder and cross-attention mechanism, is trained to integrate the global style cues from the exemplar image.
Throughout this phase, the Exemplar Network is not engaged, and pre-trained parameters of VAE and Global Encoder are kept frozen.
The primary goal of this stage is to learn the model to translate from the control into high-quality images that simultaneously preserve the structure of the target control and embody the coarse style of the exemplar.
This is achieved through a reconstruction manner, wherein the target image is synthesized using its control and the target image itself as the exemplar.

In the second stage, the Exemplar Network and Exemplar Attention Module are incorporated into previously trained denoising U-Net.
It enables focused training of the Exemplar Network and the Exempler Attention Module within the denoising U-Net, while the other parameters of the network are kept frozen.
The overall training is conducted following the strategy outlined in \cite{cocosnet}, which employs the predefined exemplar and target pairs.
This strategy facilitates a concentrated learning process while emphasizing the detailed integration of the exemplar style and specific characteristics of the target.

\subsection{Sampling Strategy}\label{sec:sampling}
The inference process is similar to BBDM~\cite{bbdm} that employs the deterministic ODE sampler~\cite{ddim}. 
Given a inference timesteps $\{t'_s\}_{s=1}^S\sim[1:T]$, the sampling process is formulated as:
\begin{equation}
\boldsymbol{x}_{t'_{s-1}}=c_{x {t'_s}} \boldsymbol{x}_{t'_s}+c_{y {t'_s}} \boldsymbol{x}_{T}-c_{\epsilon {t'_s}} \epsilon_\theta\left(\boldsymbol{x}_{t'_s}, \tau_\theta(\boldsymbol{I}_\mathcal{Y}), \psi_\theta(\boldsymbol{x}_\mathcal{Y},\tau_\theta(\boldsymbol{z}_\mathcal{Y})), {t'_s}\right)+\sqrt{\tilde{\delta}_{t'_s}} \boldsymbol{\epsilon}
\end{equation}
where $c_{\epsilon xt}, c_{\epsilon yt}, c_{\epsilon t}$ are weighting coefficients for each terms.
The whole training process and sampling process are summarized in Alg.~\ref{alg:training} and~\ref{alg:sampling}.
%
%
%
%
% ---------------------------------------------------------------
% Experiment 
% ---------------------------------------------------------------
\section{Experiments}\label{sec:experiments}
In this section, we present the experimental results of the proposed method.
We conduct three tasks to evaluate our model: Edge-to-photo, mask-to-photo, and pose-to-photo.
We perform extensive ablation studies to analyze the effect of each essential component of the proposed method.
Also, we provide qualitative and quantitative comparisons with state-of-the-art methods.
Implementation details and detailed architecture are described in supplementary material.

\noindent\textbf{Datasets. }
We conduct three tasks to evaluate our model: Edge-to-photo, mask-to-photo, and pose-to-photo.
For mask-guided and edge-guided image generation tasks, the CelebA-HQ~\cite{celebahq} dataset is used and we construct the edge maps using the Canny edge detector following~\cite{cocosnet,cocosnetv2}.
For the pose-guided image generation task, we use deepfashion~\cite{deepfashion} dataset that consists of $52,712$ images with a keypoints annotation.
For all tasks, the split of train and validation pairs is consistent with CoCosNet~\cite{cocosnet} policies. 
% ====================================================================================
\input{tab/quantitative}
% ====================================================================================
\subsection{Qualitative Evaluation}
We present a comparison of qualitative results (~\cref{fig:qualitative}) with existing methods~\cite{cocosnet, unite, dynast} at three tasks.
The results demonstrate that our method effectively transfers the detailed texture from the exemplar to the target, concurrently preserving the structure of controls.
Notably, in pose-to-photo, our approach exhibits superiority in capturing detailed patterns and minor objects, such as a cap, which other methods often overlook due to the limitation of matching frameworks.
These advantages show the capability of our proposed method that fully leverages the diffusion framework that ensures a more holistic and precise depiction.
On the other hand, in edge-to-photo and mask-to-photo tasks, while existing methods also achieve photo-realism, they often tend to overfit to the ground truth (\eg  UNITE~\cite{unite}), thereby constraining its generality. 
However, our method not only accurately transposes the texture of the exemplar but also adeptly conserves the structure.
Moreover, the images synthesized through our method demonstrably excel in photo-realistic attributes against other methods.
% ====================================================================================
\input{tab/consistency}
% ====================================================================================
\subsection{Quantitative Evaluation}
\noindent\textbf{Evaluation Metrics. }
We report the Fréchet Inception Distance (FID)~\cite{fid} and Sliced Wasserstein Distance (SWD)~\cite{swd} metrics to evaluate the image perceptual quality by reflecting the distance of feature distributions between real images and generated samples.
And we also measure LPIPS~\cite{lpips} to evaluate the diversity of translated images.
On the other hand, we show the semantic, color, and texture consistency in \cref{tab:consistency}, also under the same setting as~\cite{cocosnet}. 

\noindent\textbf{Image Quality. }
~\cref{tab:quantitative} presents a quantitative evaluation against state-of-the-art matching-based methods~\cite{cocosnet, cocosnetv2, unite, dynast, midms}, showing that our method is competitive both on image quality and diversity across various tasks.
Additionally, in the mask-to-photo task, our method demonstrates superior performance, whereas matching-based methods struggle due to their reliance on cross-domain matching—a notably arduous endeavor when masks offer scant correspondence cues. 
Conversely, by leveraging diffusion models, our method iteratively translates images from masks via noise prediction.
This enables our approach to excel in scenarios with limited direct correspondences, showcasing its robustness and adaptability.

\noindent\textbf{Consistency. }
The semantic and style consistency analysis (~\cref{tab:consistency}) evidences that our method either leads or remains competitive in style relevance scores, encompassing color and texture dimensions. 
In the pose-to-photo domain, despite achieving scores comparable to other methods~\cite{dynast, cocosnet}, a visual assessment (~\cref{fig:qualitative}) reveals our method's distinct proficiency in retaining intricate details such as patterns or textures. 
This achievement is attributable to our integrated framework, which combines the Exemplar Network and Global Encoder within a Brownian bridge diffusion model construct. 
As a result, our methodology not only yields photo-realistic images but also ensures the preservation of texture and style congruence with the exemplar input, underscoring its effectiveness in generating visually coherent outputs.

\subsection{Comparison to State-of-the-Arts Diffusion Methods}%\vspace{-5pt}
We compare our framework against prevalent state-of-the-art (SOTA) diffusion-based techniques, as shown in ~\cref{fig:ablation-vs-diffusion} and ~\cref{tab:ablation-vs-diffusion}. 
Based on the Stable Diffusion framework~\cite{ldm}, we incorporate the ControlNet~\cite{controlnet} and IP-Adapter~\cite{ipadapter} to facilitate structured and stylistic control, respectively. 
While the existing SOTA method adeptly captures the control structure and generates photo-realistic images, our method more accurately reflects the style of the exemplar.
Notably, diffusion-based approaches, conditioned on multiple information including those derived from ControlNet, textual prompts, and image embeddings, tend to be overly sensitive to hyperparameters such as control and embedding guidance scales.
Conversely, our model, predicated on a Brownian Bridge diffusion process and exclusively conditioned on the exemplar, assures a more effective generation process. 
Moreover, the capacity of the existing methods for transferring the finer details in exemplar is somewhat constrained by their reliance on CLIP embeddings which often overlook small details.
In contrast, our framework, underpinned by the Exemplar Network and Exemplar Attention Module, 
demonstrates superior adeptness in transposing textures from the exemplar. 

% ====================================================================================
\subsection{Ablation Study}
To validate the efficacy of our proposed architecture, we conduct ablation studies focusing on the following configurations: (1) omitting the Global Encoder, (2) utilizing the baseline model~\cite{bbdm} integrated with \texttt{CLIP}, (3) implementing \texttt{DINOv2}, and (4) employing our complete architecture on the edge-to-photo translation task.
As illustrated in ~\cref{fig:vsclip}, our findings reveal that the \texttt{DINOv2}-based Global Encoder surpasses the \texttt{CLIP} in generating images with higher detail fidelity. 
While \texttt{CLIP} effectively captures the general characteristics of the reference image, ensuring a level of resemblance, it does not fully encapsulate the intricacies of the details. 
Additionally, with our Exemplar Network, inputs with spatial misaligned control and exemplar often result in the generation of "blurry" images when relying exclusively on features of Global Encoder.
In contrast, our complete framework demonstrates superior performance across all assessed dimensions, highlighting its architectural advantage.
Quantitative assessments further underscore the importance of our design choices, as detailed in ~\cref{tab:vsclip}. 
% ====================================================================================

% ---------------------------------------------------------------
% Conclusion
% ---------------------------------------------------------------
\section{Conclusion} \label{conclusion}
In this study, we presented EBDM, a novel stochastic Brownian bridge diffusion-based approach for exemplar-guided image translation.
Our method is structured around three important components: denoising U-Net equipped with Exemplar Attention Module, Global Encoder, and Exemplar Network. 
By leveraging the Brownian Bridge framework, which translates from fixed data points as structural control to photo-realistic images, our method is exclusively conditioned to the style information, thereby the framework more robust and stable.

Additionally, we propose the Exemplar Network and Exemplar Attention Module to selectively incorporate the style information from exemplar images into the denoising process.
Our method not only stands competitive or surpasses existing methods across the three distinct tasks.
Furthermore, our methods also achieve a significant improvement in visual results not only in photorealism but also in the precise transfer of fine details such as patterns and accessories present in the exemplar images.
% ---------------------------------------------------------------
% Acknowledgement
% ---------------------------------------------------------------
\section*{Acknowledgements}
This research was supported by the National Research Foundation of Korea (NRF) grant funded by the Korea government (MSIP) (NRF2021R1A2C2006703) and the Yonsei Signature Research Cluster Program of 2024 (2024-22-0161).
\clearpage
\bibliographystyle{splncs04}
\bibliography{main.bib}
\clearpage
% supplementarty
\input{main_supp}
\end{document}

%% file: tab/algorithm.tex
\begin{algorithm}[bt]
\caption{Training}\label{alg:training}
\begin{algorithmic}[1]
    \Repeat
    \State $(\boldsymbol{x}_T,\boldsymbol{x}_0) \sim q_{\text{data}}(\mathcal{X},\mathcal{Y})$
        \Comment{Sample paired data}
    \State $\boldsymbol{I}_\mathcal{Y} \sim q_{\text{data}}(\mathcal{Y})$
    \Comment{Sample exemplar}
    \State $t \sim \text{Uniform}(1, \ldots, T)$
    \Comment{diffusion timesteps}
    \State $t_{ref} \gets 0$ \Comment{reference timestep}
    \State $\boldsymbol{\epsilon} \sim \mathcal{N}(\boldsymbol{0},\boldsymbol{I})$\Comment{sample Gaussian noise}
    \State $\boldsymbol{G} \gets \tau_\theta(\boldsymbol{I}_\mathcal{Y})$
    \Comment{Forward pass through Global Encoder}
    \State $\boldsymbol{F} \gets \psi_\theta(\boldsymbol{x}_\mathcal{Y}, t_{ref}, \boldsymbol{G})$
    \Comment{Forward pass through Exemplar Network}
    \State $\boldsymbol{x}_t \gets \left(1-m_t\right) \boldsymbol{x}_0+m_t
    \boldsymbol{y}+\sqrt{\delta_t} \boldsymbol{\epsilon}$
    \Comment{Forward bridge diffusion process}
    \State $\nabla_\theta\left\|m_t\left(\boldsymbol{y}-\boldsymbol{x}_0\right)+\sqrt{\delta_t} \boldsymbol{\epsilon}-\boldsymbol{\epsilon}_\theta\left(\boldsymbol{x}_t, \boldsymbol{G}, \boldsymbol{F}, t\right)\right\|^2$
    \Comment{Gradient descent step}
    \Until{converged}
    \end{algorithmic}
    \end{algorithm}
    \begin{algorithm}[bt]
        \caption{Sampling}\label{alg:sampling}
        \begin{algorithmic}[1]
        \State $\boldsymbol{x}_T\sim q_{\text{data}}(\mathcal{X})$
        \Comment{Sample control input}
        \State $\boldsymbol{I}_\mathcal{Y} \sim q_{\text{data}}(\mathcal{Y})$
        \Comment{Sample exemplar input}
        \State $t'_s\leftarrow\{t'_S,\cdots t'_1\}\sim \{t_T,\cdots,t_1\}$
        \Comment{$S$ Inference timesteps}
        \State $\boldsymbol{G} \gets \tau_\theta(\boldsymbol{I}_\mathcal{Y})$
    \Comment{Forward pass through Global Encoder}
    \State $\boldsymbol{F} \gets \psi_\theta(\boldsymbol{x}_\mathcal{Y}, \boldsymbol{G})$
    \Comment{Forward pass through Exemplar Network}
        \For{$s=S, \ldots, 1$}
    % \If{$s = N $}
    \State $\boldsymbol{\epsilon} \sim \mathcal{N}(\mathbf{0}, \mathbf{I})$ if $s>1$, else $\boldsymbol{\epsilon}=0$
    \State $\boldsymbol{x}_{t'_{s-1}}=c_{x {t'_s}} \boldsymbol{x}_{t'_s}+c_{y {t'_s}} \boldsymbol{x}_T-c_{\boldsymbol{\epsilon}{t'_s}} \boldsymbol{\epsilon}_\theta\left(\boldsymbol{x}_{t'_s}, \boldsymbol{G}, \boldsymbol{F}, {t'_s}\right)+\sqrt{\tilde{\delta}_{t'_s}} \boldsymbol{\epsilon}$
    \Comment{Take sampling step}
    \EndFor
    \Return $\boldsymbol{x}_0$ %$$\mathcal{D}(\boldsymbol{x}_0)$
    % \Comment{Decode latent to image space}
    \end{algorithmic}
\end{algorithm}

%% file: tab/quantitative.tex
\begin{table*}[bt]
\resizebox{\textwidth}{!}{
      \centering
      \begin{tabular}{l ccc m{.1cm} ccc m{.1cm} ccc}
      \hline
      \multirow{2}{*}{\textbf{Method}} &
      \multicolumn{3}{c}{\textbf{DeepFashion}} &&
      \multicolumn{3}{c}{\textbf{CelebA-HQ (Edge)}} &&
      \multicolumn{3}{c}{\textbf{CelebA-HQ (Mask)}} \\
      \cline { 2 - 4 } \cline { 6 - 8 }  \cline {10-12}
      &
      FID$\downarrow$&SWD$\downarrow$&LPIPS$\uparrow$&&
      FID$\downarrow$&SWD$\downarrow$&LPIPS$\uparrow$&&
      FID$\downarrow$&SWD$\downarrow$&LPIPS$\uparrow$\\
      \hline
      \textbf{Pix2PixHD}~\cite{pix2pixhd}&
      25.20     &16.40	&\texttt{N/A}   &&
      42.70     &33.30	&\texttt{N/A}   &&
      43.69     &34.82      &\texttt{N/A}    \\

      \textbf{SPADE}~\cite{spade}&
      36.20	&27.80	&0.231 &&
      31.50	&26.90	&0.187 &&
      39.17     &29.78  &0.254  \\

      \textbf{SelectionGAN}~\cite{selectiongan}&
      38.31	&28.21	&0.223 &&
      34.67	&27.34	&0.191 &&
      42.41     &30.32  &0.277  \\

      \textbf{SMIS}~\cite{smis}&
      22.23	&23.73	&0.240 &&
      23.71	&22.23	&0.201 &&
      28.21     &24.65      &0.301  \\

      \textbf{SEAN}~\cite{sean}&
      16.28	&17.52	&0.251 &&
      18.88	&19.94	&0.203 &&
      17.66     &14.13      &0.285\\

      \textbf{CoCosNet}~\cite{cocosnet}&
      14.40	&17.20	&0.272 &&
      14.30	&15.30	&0.208 &&
      21.83     &12.13      &0.292  \\

      \textbf{CoCosNetv2}~\cite{cocosnetv2}&
      12.81	&16.53	&0.283 &&
      12.85	&14.62	&0.218 &&
      20.64     &11.21      &0.303  \\

      \textbf{UNITE}~\cite{unite}&
      13.08	&16.65	&0.278 &&
      13.15	&14.91	&0.213 &&
      \texttt{N/A}       &\texttt{N/A}        &\texttt{N/A}   \\

      \textbf{RABIT}~\cite{rabit}&
      12.58     &16.03      &\underline{0.284}  &&
      \textbf{11.67}     &14.22      &0.219  &&
      20.44     &\textbf{11.18}      &0.307  \\

      \textbf{MCL-Net}~\cite{mclnet}&
      12.89	&16.24	&\textbf{0.286}  &&
      12.52	&14.21	&0.216  &&
      \texttt{N/A}       &\texttt{N/A}        &\texttt{N/A}    \\
      \textbf{MIDMs}~\cite{midms}&
      \underline{10.89}	&\textbf{10.10}	&0.279  &&
      15.67   &\textbf{12.34}	&\underline{0.224}  &&
      \texttt{N/A}       &\texttt{N/A}        &\texttt{N/A}    \\
      \hline
      \textbf{Ours} &
      \textbf{10.62}      &\underline{12.40}  &0.255   && % deepfashion
      \underline{11.84}     &\underline{12.10}   &\textbf{0.227}   && % celeb edge
      \textbf{12.21}     & \underline{11.34}    &\textbf{0.215}   \\ % celeb mask
      \hline
      % \bottomrule
      \end{tabular}
      } 
    \vspace{5pt}
      \caption{
      \textbf{Quantitative Results} in image quality. 
      Comparing our methods with state-of-the-art exemplar-guided image translation methods.
      }
  \label{tab:quantitative}
  \vspace{-20pt}
\end{table*}

%% file: tab/consistency.tex
\begin{table}[bt!]
    \centering
    \begin{tabular}{c ccc m{.1cm} ccc}%$ m{.1cm} ccc}
        \hline \multirow{2}{*}{\textbf{Method}} &
        \multicolumn{3}{c}{\textbf{DeepFashion}} & &
        \multicolumn{3}{c}{\textbf{CelebA-HQ (Edge)}} \\ %& &
        % \multicolumn{3}{c}{\textbf{CelebA-HQ (Mask)}} \\

        \cline { 2 - 4 } \cline { 6 - 8 } &
        Sem. $\uparrow$ & Col. $\uparrow$ & Tex. $\uparrow$ &&
        % Sem. $\uparrow$ & Col. $\uparrow$ & Tex. $\uparrow$ &&
        Sem. $\uparrow$ & Col. $\uparrow$ & Tex. $\uparrow$ \\
        \hline
        \textbf{Pix2PixHD}~\cite{pix2pixhd} &
        0.943 & \texttt{N/A} & \texttt{N/A} && 0.914 & \texttt{N/A} & \texttt{N/A} \\% && - & - & - \\

        \textbf{SPADE}~\cite{spade} &
        0.936 & 0.943 & 0.904 && 0.922 & 0.955 & 0.927 \\% && - & - & - \\

        \textbf{MUNIT}~\cite{munit} &
        0.910 & 0.893 & 0.861 && 0.848 & 0.939 & 0.884 \\%  && - & - & - \\

        \textbf{EGSC-IT}~\cite{egsc-it} &
        0.942 & 0.945 & 0.916 && 0.915 & 0.965 & 0.942 \\%  && - & - & - \\

        \textbf{CoCosNet}~\cite{cocosnet} &
        0.968 & 0.982 & \textbf{0.958} && \underline{0.949} & 0.977 & 0.958 \\%  && - & - & - \\

        \textbf{CoCosNet-v2}~\cite{cocosnetv2} &
        \underline{0.969} & 0.974 & 0.925 && 0.948 & 0.975 & 0.954  \\% && - & - & - \\

        \textbf{UNITE}~\cite{unite} &
        0.957 & 0.973 & 0.930 && \textbf{0.952} & 0.966 & 0.950 \\%  && - & - & - \\
        
        \textbf{DynaST}~\cite{dynast} &
        \textbf{0.975} & \underline{0.974} & 0.937 && \textbf{0.952} & 0.980 & \textbf{0.969} \\%  && - & - & - \\
        \textbf{MIDMs}~\cite{midms} &
        \texttt{N/A} & \texttt{N/A} & \texttt{N/A} && 0.915 & 0.982 & 0.962 \\%  && - & - & - \\
        \textbf{MATEBIT}~\cite{matebit} &
        \texttt{N/A} & \texttt{N/A} & \texttt{N/A} && \underline{0.949} & \textbf{0.986} & \underline{0.966} \\%  && - & - & - \\
        \hline
        \textbf{Ours} &
        0.932 & \textbf{0.982} & \underline{0.939} && 0.920 & \underline{0.984} & 0.968 \\%  && - & - & - \\
        \hline
    \end{tabular}
    \vspace{5pt}
    \caption{Quantitative metrics of semantic (Sem.), color (Col.), and texture (Tex.) consistency on two datasets compared with state-of-the-art image synthesis methods.}
    \label{tab:consistency}
    % \vspace{-20pt}
\end{table}

%% file: main_supp.tex
% \documentclass[runningheads]{llncs}
% % TODO FINAL: choose one
% \usepackage{eccv}
% % \usepackage[mobile]{eccv}

% % -------------------------------------------------
% % Packages
% \usepackage{eccvabbrv}
% \usepackage{graphicx}
% \usepackage{booktabs}
% \usepackage[accsupp]{axessibility}

% % FOR REVIEW
% % TODO FINAL:
% \usepackage[pagebackref,breaklinks,colorlinks]{hyperref}

% \usepackage{orcidlink}
% \usepackage{array}
% \usepackage{makecell}
% \usepackage{amsmath}
% \usepackage[makeroom]{cancel}

% \usepackage{amssymb}
% \usepackage{booktabs}
% \usepackage{multirow}
% \usepackage{tabularx}
% \usepackage{adjustbox}
% \usepackage{float}
% \usepackage{xcolor,subcaption}
% \usepackage{placeonpage}
% \documentclass[runningheads]{llncs}
% % -------------------------------------------------
% % ECCV Packages
% \usepackage{eccv}
% % \usepackage[mobile]{eccv}
% \usepackage{eccvabbrv}
% \usepackage{graphicx}
% \usepackage{booktabs}
% \usepackage[accsupp]{axessibility} % Improves PDF readability for those with disabilities.
% \usepackage{hyperref}
% \usepackage{orcidlink}

% % Custom Packages
% \usepackage{tabularx}
% \usepackage[export]{adjustbox}
% \usepackage{varwidth}
% \usepackage{array}
% \usepackage{makecell}
% \usepackage{amsmath}
% \usepackage{amssymb}
% \usepackage{multirow}
% \usepackage{float}
% \usepackage{xcolor,subcaption}
% \usepackage{placeonpage}
% \usepackage{algorithm}
% \usepackage[noend]{algpseudocode}

% % -------------------------------------------------
% \begin{document}
\title{Supplementary Materials for \\ EBDM: Exemplar-guided Image Translation with Brownian-bridge Diffusion Models.}
\titlerunning{Supplementary for EBDM}
\author{
  Eungbean Lee\inst{1}\orcidlink{0000-0003-4839-8540}\and
  Somi Jeong\inst{2}\orcidlink{0000-0002-0906-0988}\and
  Kwanghoon Sohn\inst{1,3}%\orcidlink{0000-0002-3715-0331}
}
\authorrunning{E.~Lee et al.}

\institute{
Yonsei University, Seoul, Korea \\\email{\{eungbean,khsohn\}@yonsei.ac.kr}\and
NAVER LABS \\\email{somi.jeong@naverlabs.com} \and
Korea Institute of Science and Technology (KIST), South Korea}
\maketitle
\setcounter{equation}{9}
\setcounter{figure}{6}
\setcounter{table}{4}

This supplementary material provides details that are not included in the main paper due to space limitations.
We provide the explanation of deduction details at~\cref{sec:bbdm} and advantages over DDPMs~\cref{sec:advantages}.
Then the implementation details of EBDM will be presented at~\cref{sec:experiment-supp}.
Finally, we will present more qualitative experiment results.

% ================================================================================================
% Brownian Bridge
% ================================================================================================
\section{Brownian Bridge Diffusion Models}\label{sec:bbdm}
In this section, we provide more details of Brownian Bridge Diffusion Models (BBDM)~\cite{bbdm}.
The BBDM aims to connect two image domains via discrete Brownian bridges.
Assuming that the start point and end point of the diffusion process,
$(\boldsymbol{x}_0,\boldsymbol{x}_T)=(\boldsymbol{x},\boldsymbol{y})\sim q_{data}(\boldsymbol{x},\boldsymbol{y})$,
BBDM learns to approximately sample from $q_{data}(\boldsymbol{x}|\boldsymbol{y})$ by reversing the diffusion bridge with boundary distribution
$q_{data}(\boldsymbol{x},\boldsymbol{y})$, given a training set of paired samples drawn from $q_{data}(\boldsymbol{x},\boldsymbol{y})$.

% ================================================================================================
% Forward Process
% ================================================================================================
\subsection{Forward Process}
Given initial state $\boldsymbol{x}_0$ and destination state $\boldsymbol{y}$,
the forward diffusion process of the Brownian Bridge can be defined as:
\begin{equation}\begin{gathered}\label{eq:bb2}
    p\left(\boldsymbol{x}_t \mid \boldsymbol{x}_0, \boldsymbol{x}_T\right)\
    =\mathcal{N}\left(\boldsymbol{x}_t;(1-m_t)\boldsymbol{x}_0 + m_t \boldsymbol{y}, \boldsymbol{\delta}_t \boldsymbol{I}\right)\\
    \text{where} \
    \quad m_t=\frac{t}{T},\quad \
    % \delta_t=\frac{t(T-t)}{T} \red{=\frac{m_t(1-m_t)}{T}}
    \delta_t = 2s(m_t-m^2_t) %\red{=2s\frac{t}{T}\left(1-\frac{t}{T}\right)}
\end{gathered}\end{equation}
where $T$ is the total steps of the diffusion process, $s$ is the variance factor, and $\delta_t$ is the variance that is designed to preserve the maximum at $t=2/T$ as identity, \ie $\delta_{max}=\frac{1}{2}$. The variance factor $s$ scales the maximum variance to control diffusion diversity, and we set $s=1$ as the default. 
The intermediate state $\boldsymbol{x}_t$ in its discrete form can be determined by calculating:

\begin{equation}\label{eq:fp1}
\boldsymbol{x}_t=\left(1-m_t\right) \boldsymbol{x}_0+m_t \boldsymbol{y}+\sqrt{\delta_t} \boldsymbol{\epsilon}_t\quad\text{where}\quad\boldsymbol{\epsilon}_t\sim\mathcal{N}(\boldsymbol{0},\boldsymbol{I})
\end{equation}
We can express $\boldsymbol{x}_0$ with $\boldsymbol{x}_{t}$ and \cref{eq:fp1}:
\begin{equation}\label{eq:fp2}
    \boldsymbol{x}_0 = \frac{1}{1-m_t}\left(\boldsymbol{x}_t - m_t\boldsymbol{y} - \sqrt{\delta_t} \boldsymbol{\epsilon}_t\right)
\end{equation}

Thus, the transition probability $q\left(\boldsymbol{x}_t \mid \boldsymbol{x}_{t-1}, \boldsymbol{y}\right)$
can be derived by substituting the expression of~\cref{eq:fp1} and~\cref{eq:fp2}:

\begin{equation}\begin{aligned}\label{eq:fp-bb}
    q\left(\boldsymbol{x}_t \mid \boldsymbol{x}_{t-1}, \boldsymbol{y}\right) \
    = \mathcal{N}(\boldsymbol{x}_t ; \
        \hat{\mu}_t(\boldsymbol{x}_{t-1},\boldsymbol{y}) \
        % , \hat{\delta}_t \boldsymbol{I})
        , \hat{\delta}_t  \boldsymbol{I})
\end{aligned}\end{equation}

\begin{equation}\begin{aligned}\label{eq:fp-bb2}
    \text{where},\quad \
    \hat{\mu}_t(\boldsymbol{x}_{t-1},\boldsymbol{y}) &= \
        \frac{1-m_t}{1-m_{t-1}} \boldsymbol{x}_{t-1} \
        +\left(m_t-\frac{1-m_t}{1-m_{t-1}} m_{t-1}\right) \boldsymbol{y} \\
        % \hat{\delta}_t &=
        \hat{\delta}_t &= \delta_{t \mid t-1} =
        \delta_t - \delta_{t-1}\frac{\left(1-m_t\right)^2}{\left( 1-m_{t-1}\right)^2}
\end{aligned}\end{equation}

% ================================================================================================
% Reverse Process
% ================================================================================================
\subsection{Reverse Process}
The reverse process of BBDM is to predict $\boldsymbol{x}_{t-1}$ given $\boldsymbol{x}_t$:

\begin{equation}\label{eq:rp1}
p_\theta\left(\boldsymbol{x}_{t-1} \mid \boldsymbol{x}_t, \boldsymbol{y}\right) \
=\mathcal{N}\left(\boldsymbol{x}_{t-1} ; \
    \boldsymbol{\mu}_\theta\left(\boldsymbol{x}_t, t\right), \
    \tilde{\delta}_t \boldsymbol{I}\right)
\end{equation}
where $\boldsymbol{\mu}_\theta\left(\boldsymbol{x}_t, t\right)$ represents the predicted mean, and $\tilde{\delta}_t$ denotes the variance of the noise at each step.

% ================================================================================================
% Training Objectives
% ================================================================================================
\subsection{Training Objectives}
The training procedure involves optimizing the Evidence Lower Bound (ELBO) for the Brownian Bridge diffusion process, which is expressed as:
\begin{equation}\begin{aligned}\label{eq:elbo-supp}
E L B O =&-\mathbb{E}_q \big(\
\cancel{D_{K L}\left(q\left(\boldsymbol{x}_T \mid \boldsymbol{x}_0, \boldsymbol{y}\right) \| p\left(\boldsymbol{x}_T \mid \boldsymbol{y}\right)\right)} \qquad \because \boldsymbol{x}_T = \boldsymbol{y} \\
& +\sum_{t=2}^T D_{KL} \left( q\left(\boldsymbol{x}_{t-1} \mid \boldsymbol{x}_t, \boldsymbol{x}_0, \boldsymbol{y}\right) \| p_\theta\left(\boldsymbol{x}_{t-1} \mid \boldsymbol{x}_t, \boldsymbol{y}\right)\right) \\
& -\log p_\theta\left(\boldsymbol{x}_0 \mid \boldsymbol{x}_1, \boldsymbol{y}\right) \big)
\end{aligned}\end{equation}

By combining~\cref{eq:fp-bb} and~\cref{eq:fp-bb2},
the formula $q\left(\boldsymbol{x}_{t-1} \mid \boldsymbol{x}_t, \boldsymbol{x}_0, \boldsymbol{y}\right)$ in the second term can be derived from Bayes' theorem and the Markov chain property:
\begin{equation}\begin{aligned}
q\left(\boldsymbol{x}_{t-1} \mid \boldsymbol{x}_t, \boldsymbol{x}_0, \boldsymbol{y}\right) & =\frac{q\left(\boldsymbol{x}_t \mid \boldsymbol{x}_{t-1}, \boldsymbol{y}\right) q\left(\boldsymbol{x}_{t-1} \mid \boldsymbol{x}_0, \boldsymbol{y}\right)}{q\left(\boldsymbol{x}_t \mid \boldsymbol{x}_0, \boldsymbol{y}\right)} \\
& =\mathcal{N}\left(\boldsymbol{x}_{t-1} ; \tilde{\boldsymbol{\mu}}_t\left(\boldsymbol{x}_t, \boldsymbol{x}_0, \boldsymbol{y}\right), \tilde{\delta}_t \boldsymbol{I}\right)
\end{aligned}\end{equation}

%%%%%% \mu (x_t, x_0, y) %%%%%%
% where the mean value term is:
% \begin{equation}\begin{aligned}
% \tilde{\boldsymbol{\mu}}_t\left(\boldsymbol{x}_t, \boldsymbol{x}_0, \boldsymbol{y}\right) \
%  & =\frac{\delta_{t-1}}{\delta_t} \frac{1-m_t}{1-m_{t-1}} \boldsymbol{x}_t \\
%  & +\left(1-m_{t-1} \frac{\hat{\delta}_t}{\delta_t}\right) \boldsymbol{x}_0 \\
%  & +\left(m_{t-1}-m_t \frac{1-m_t}{1-m_{t-1}} \frac{\delta_{t-1}}{\delta_t}\right) \boldsymbol{y}
% \end{aligned}\end{equation}

The mean value term $\tilde{\boldsymbol{\mu}}_t\left(\boldsymbol{x}_t, \boldsymbol{x}_0, \boldsymbol{y}\right)$ can be reformulated as
$\tilde{\boldsymbol{\mu}}_t\left(\boldsymbol{x}_0, \boldsymbol{y}\right)$ by utilizing reparameterization method~\cite{ddpm}:

\begin{equation}
    \begin{aligned}\label{eq:mean}
    \tilde{\boldsymbol{\mu}}_t\left(\boldsymbol{x}_t, \boldsymbol{y}\right)&=c_{x t} \boldsymbol{x}_t+c_{y t} \boldsymbol{y}+c_{\epsilon t}\left(m_t\left(\boldsymbol{y}-\boldsymbol{x}_0\right)+\sqrt{\delta_t} \boldsymbol{\epsilon}\right) \\
    \text{where,} \quad \
    c_{x t}&=\frac{\delta_{t-1}}{\delta_t} \frac{1-m_t}{1-m_{t-1}}+\frac{\hat{\delta}_t}{\delta_t}\left(1-m_{t-1}\right) \\
    c_{y t}&=m_{t-1}-m_t \frac{1-m_t}{1-m_{t-1}} \frac{\delta_{t-1}}{\delta_t} \\
    c_{\epsilon t}&=\left(1-m_{t-1}\right) \frac{\hat{\delta}_t}{\delta_t}
    \end{aligned}
\end{equation}
And the variance term is:
\begin{equation}
    \tilde{\delta}_t=\frac{\hat{\delta}_t \cdot \delta_{t-1}}{\delta_t}% = \frac{2(t-1)(T-t+1)^2}{tT}
\end{equation}

As the neural network $\boldsymbol{\epsilon}_\theta$ predict the noise, thus, the reverse process~\cref{eq:rp1} can be reformulated as:
\begin{equation}
    \boldsymbol{\mu}_{\boldsymbol{\theta}}\left(\boldsymbol{x}_t, \boldsymbol{y}, t\right)=c_{x t} \boldsymbol{x}_t+c_{y t} \boldsymbol{y}+c_{\epsilon t} \boldsymbol{\epsilon}_\theta\left(\boldsymbol{x}_t, t\right)
\end{equation}

Therefore, the training objective ELBO in~\cref{eq:elbo-supp} can be simplified as:
\begin{equation}
    \mathbb{E}_{\boldsymbol{x}_0, \boldsymbol{y}, \boldsymbol{\epsilon}}\left[c_{\epsilon t}\left\|m_t\left(\boldsymbol{y}-\boldsymbol{x}_0\right)+\sqrt{\delta_t} \boldsymbol{\epsilon}-\boldsymbol{\epsilon}_\theta\left(\boldsymbol{x}_t, t\right)\right\|^2\right]
\end{equation}
The weighting function $c_{x t}, c_{y t}, c_{\epsilon t}$ in~\cref{eq:mean} is used in Eq. (7) and (8).

% ================================================================================================
% Experiments
% ================================================================================================
\clearpage
\section{Advantages over DDPMs.}\label{sec:advantages}
The primary motivation for choosing BBDMs is to 1) reduce the number of conditions and 2) utilize an end-to-end training framework.

\noindent\textbf{1) Reducing the number of conditions.}
The primary motivation for choosing the Brownian Bridge is to simplify the conditioning mechanism.
By reducing the number of conditions,  
thereby minimizing the parameters, training times, and the risk of overfitting, while enhancing robustness. 
Increasing the number of conditions $\boldsymbol{c}=\{c_1,\cdots,c_n\}$ significantly impacts both training and performance. 
The conditional distribution (Eq.~\ref{cond}), and reverse process (Eq.~\ref{eq:reverse}) can be described as:
\vspace{-12pt}
\begin{align}
P(x|\boldsymbol{c})=\frac{P(x)}{P\left(c_1, \ldots, c_n\right)}\prod_{i=1}^N P\left(c_i \mid x\right) \propto \prod_{i=1}^N\frac{ P\left(x \mid c_i\right)}{P(x)}\label{cond} \\
p_\theta\left(\boldsymbol{x}_{t-1}|\boldsymbol{x}_{t},\boldsymbol{c}):=\mathcal{N}(\boldsymbol{x}_{t-1};\mu_\theta(\boldsymbol{x}_{t},\boldsymbol{c},t),\Sigma_\theta(\boldsymbol{x}_{t},\boldsymbol{c},t)\right)\label{eq:reverse} 
\end{align}

As the number of conditions $n$ grows, the loss function becomes more complex affecting the modeling the $\mu_\theta$ and $\Sigma_\theta$.
This complexity can be quantified by the KL divergence between the true conditional distribution and the model distribution, 
indicating a more complex distribution that the model must learn to approximate accurately, leading to convergence difficulties, gradient instability, and the need for stronger regularization techniques. Simplifying the conditioning mechanism mitigates these issues by:

\begin{itemize}
\item \textbf{Reducing parameters:} Lower dimensionality in the conditional space decreases the number of parameters, 
leading the optimization landscape less complex.
\item \textbf{Reduced demand for Data Requirement:} Less data is needed to cover the distributions with the same density due to the curse of dimensionality. 
%($\text{Data\ Size} \propto 2^{\text{dims}}$).
%
\item \textbf{Less training time:} The Computational costs are reduced as the complexity of computing the gradients reduces.
% ($\text{Training Time}\propto \mathcal{O}(\text{Data Size}\times\text{Model Complexity})$).
%
\item \textbf{Lower risk of overfitting}. A simpler model is less likely to capture noise and specific characteristics of the training data, 
due to the variance of the function $\boldsymbol{\epsilon}_\theta(\boldsymbol{c},t)$ increases, 
adversely affecting generalization and stable training. 
\end{itemize}
\noindent 

\noindent\textbf{2) End-to-end training:} SD-based method takes modular approaches\footnote{\eg Stable Diffusion equipped with ControlNet and IP-Adapter.\label{sd}} that are not trained end-to-end, posing a risk of unwanted information influencing the inference. 
In contrast, our method benefits from an end-to-end training framework, enhancing integration and performance, particularly in exemplar-guided image translation tasks.

% ================================================================================================
% Experiments
% ================================================================================================
\newpage
\input{supp/tab1}

\section{More Experiment Details}\label{sec:experiment-supp}
In this section, further implementation specifics of the EBDM are elucidated, encompassing network hyperparameters (Tab.~\ref{tab:sup1}), optimization strategies, as well as computational efficiency. 

\subsection{Datasets}
We conduct three tasks to evaluate our model: Edge-to-photo, mask-to-photo, and pose-to-photo.
For mask-guided and edge-guided image generation tasks, the CelebA-HQ~\cite{celebahq} dataset is used and we construct the edge maps using the Canny edge detector following~\cite{cocosnet,cocosnetv2}.
For the pose-guided image generation task, we use deepfashion~\cite{deepfashion} dataset that consists of $52,712$ images with a keypoints annotation.
The split of train and validation pairs is consistent with CoCosNet~\cite{cocosnet} policies. 

\subsection{Training}\label{training}
All experiments are conducted utilizing a spatial resolution of $64\times64$ within the latent space.
During training, we use a batch size of 8 with gradient accumulation 2, each batch containing pairs of an input exemplar and condition following~\cite{cocosnet}.
The model is trained with \texttt{AdamW} optimizer for the learning rate of $1.0\mathrm{e}{-5}$ and learning rate decay with $\gamma={0.2}$.
The Exponential Moving Average (EMA) was adopted in the training procedure together with \texttt{ReduceLROnPlateau} learning rate scheduler.
Training is done on \texttt{Pytorch} framework and Nvidia RTX A6000 48GB GPU.

\subsection{Autoencoders}
We adopt the pretrained VQGAN presented in~\cite{ldm}, which reduces images to $64\times64$ resolution in latent space.
In edge-to-photo and mask-to-photo tasks using CelebA-HQ~\cite{celebahq}, we use VQ-regularized autoencoder with downsampling factor $f=4$ and channel dimension $3$.
For the pose-to-photo task using DeepFashion~\cite{deepfashion}, we use KL-regularized autoencoder with downsampling factor $f=8$ and channel dimension $4$.
Both the encoder and decoder are frozen during training for fair comparison.

\subsection{Computational Efficiency}
Our method improves computational cost, as demonstrated in Tab.~\ref{tab:flops}.
Our method achieves a -28.21\% reduction of FLOPs indicating faster inference time.
Furthermore, in the inference stage, the SD-based model requires extensive grid searches across the conditional parameters (\eg guidance scale, control weight, IP-adapter scale, \textit{etc}.) to achieve plausible results, which consumes significant resources.
By reducing the number of conditions, our method improves efficiency in both computational and practical uses.

\input{supp/tab4}

\subsection{Additional Qualitative Results}
Lastly, we present further qualitative results in comparison with other techniques in ~\cref{fig:sup1,fig:sup2,fig:sup3}.
Additional diverse samples with various control inputs are shown in~\cref{fig:sup1-2,fig:sup2-2,fig:sup3-2}.

% ==================================================================================================================================

\section{Limitations}
Our approach utilizes the Brownian Bridge diffusion process in latent space~\cite{ldm} to connect control and image latents effectively. 
However, the pre-trained VAE Encoder that focuses on image representation limits its ability to process control signals accurately, especially when differentiating semantically diverse elements (such as background and face in mask), focusing more on color distance rather than semantic discrepancies.

To mitigate this, prior studies~\cite{controlnet, unicontrolnet} have introduced additional control guiders. 
Yet, integrating these with the Brownian Bridge model, characterized by its reliance on two fixed endpoints, complicates the direct integration of such solutions.

\newpage\input{supp/fig1}
\newpage\input{supp/fig2}
\newpage\input{supp/fig3}
\newpage\input{supp/fig1-2}
\newpage\input{supp/fig2-2}
\newpage\input{supp/fig3-2}

% \clearpage

% \bibliographystyle{splncs04}
% \bibliography{egbib}

% \end{document}

%% file: supp/tab1.tex
\begin{table}[ht!]
% \resizebox{\textwidth}{!}{
  \centering
\begin{tabular}{ccccccc}
\hline
    model &
    z-shape &
    channels &
    \makecell{channel \\ multiplier} &
    \makecell{attention \\ resolutions} &
    \makecell{total \\ parameters} &
    \makecell{trainable \\ parameters} \\
\hline
    BBDM-f4         & $64 \times 64 \times 3$ & 128 & $1,4,8$ & $32,16,8$ & $437.81 \mathrm{M}$ & $382.49 \mathrm{M}$ \\
    Exemplar Net    & $64 \times 64 \times 3$ & 128 & $1,4,8$ & $32,16,8$ & $404.82 \mathrm{M}$ & $382.48 \mathrm{M}$ \\
    Global Encoder  & - & - & - & - & $ 86.58 \mathrm{M}$ & $0$ \\
    \hline
    EBDM-f4 & $64 \times 64 \times 3$ & 128 & $1,4,8$ & $32,16,8$  & $929.21 \mathrm{M}$ & $764.97 \mathrm{M}$ \\
\hline
\end{tabular}
\caption{Network hyperparameters for EBDM and modules.}
\label{tab:sup1}
\end{table}
\vspace{-30pt}

% [NUM PARAMS][Denoising Unet] TOTAL 382,490,243 | [TRAINABLE] 382,490,243
% [NUM PARAMS][Exemplar Net] TOTAL 382,486,528 | [TRAINABLE] 382,486,528
% [NUM PARAMS][vae.encoder] TOTAL 22,337,411 | [TRAINABLE] 0
% [NUM PARAMS][dino] TOTAL 86,580,480 | [TRAINABLE] 0

%% file: supp/tab4.tex
\begin{table*}[bt!]
\centering
\begin{tabular}{l|ccc}
\toprule
\textbf{Methods} & \textbf{FLOPs \small{(1 steps)}} &  \textbf{FLOPs \small{(50 steps)}} &\textbf{\# Parameters}\\
\midrule
SD-based   & 11.14 T    & 86.08 T & 1308.7 M \\
\midrule
\multirow{2}{*}{Ours}       & 11.37 T     & 61.72 T & 764.97 M \\
                            & \small{(+2.0\%)}  & \small{(-28.21\%)}  & \small{(-41.55\%)}\\
\bottomrule
\end{tabular}
\caption{{\textbf{Comparisions of computational costs.} Number of parameters and FLOP counts with single and 50 steps in inference.
}}
  \label{tab:flops}
\end{table*}

%% file: supp/fig1.tex
\begin{figure*}[tb]
\centering
\makebox[\textwidth]{\includegraphics[width=\textwidth]{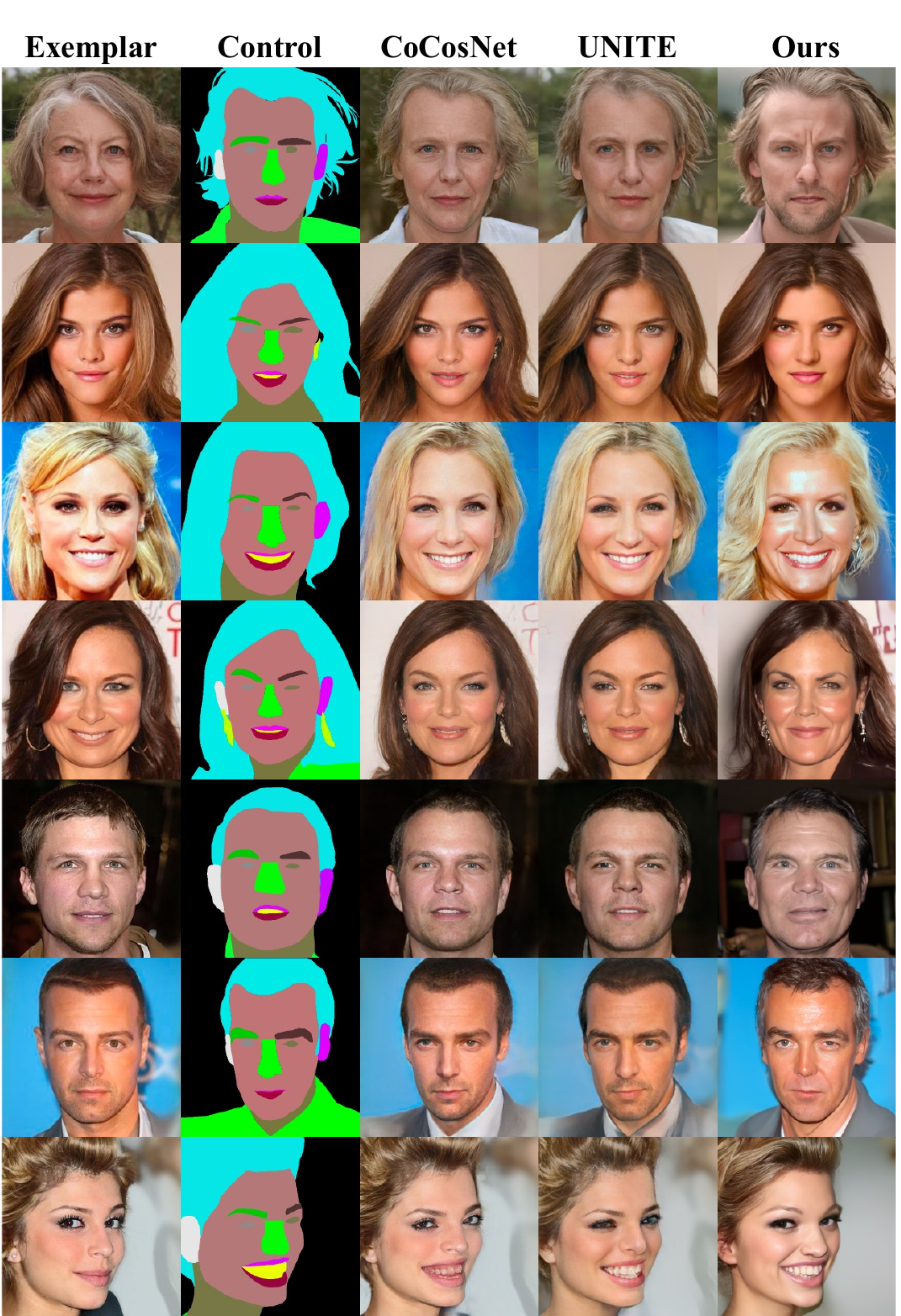}}
\caption{
    \textbf{Mask-to-image} Qualitative comparisons on the CelebAHQ-HQ Dataset.
    }
\label{fig:sup1}
\end{figure*}

%% file: supp/fig2.tex
\begin{figure*}[tb]
    \centering
    \makebox[\textwidth]{\includegraphics[width=\textwidth]{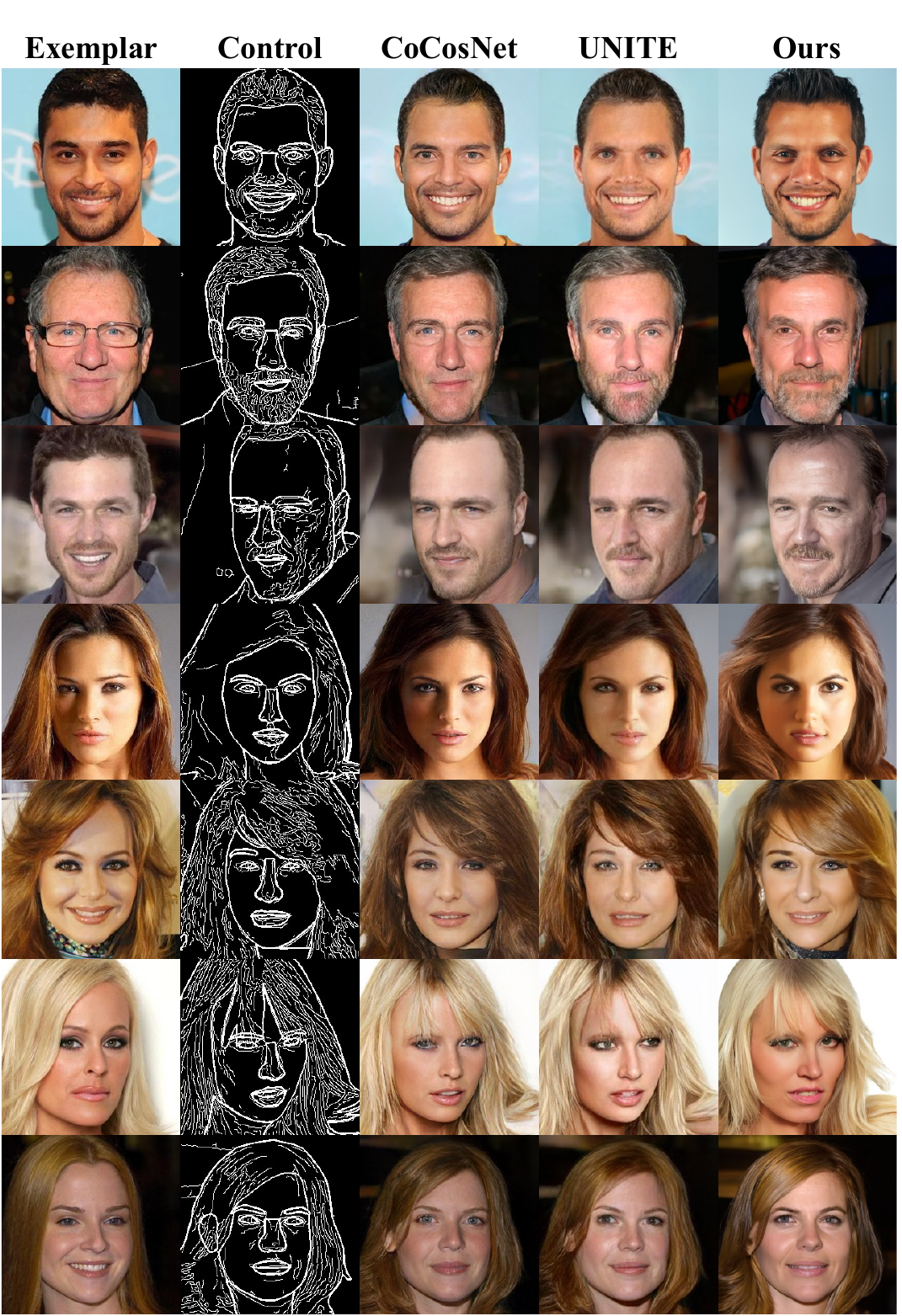}}
    \caption{\textbf{Edge-to-image} Qualitative comparisons on the CelebA-HQ Dataset.}
    \label{fig:sup2}
\end{figure*}

%% file: supp/fig3.tex
\begin{figure*}[tb]
    \centering
    \makebox[\textwidth]{\includegraphics[width=.8\textwidth]{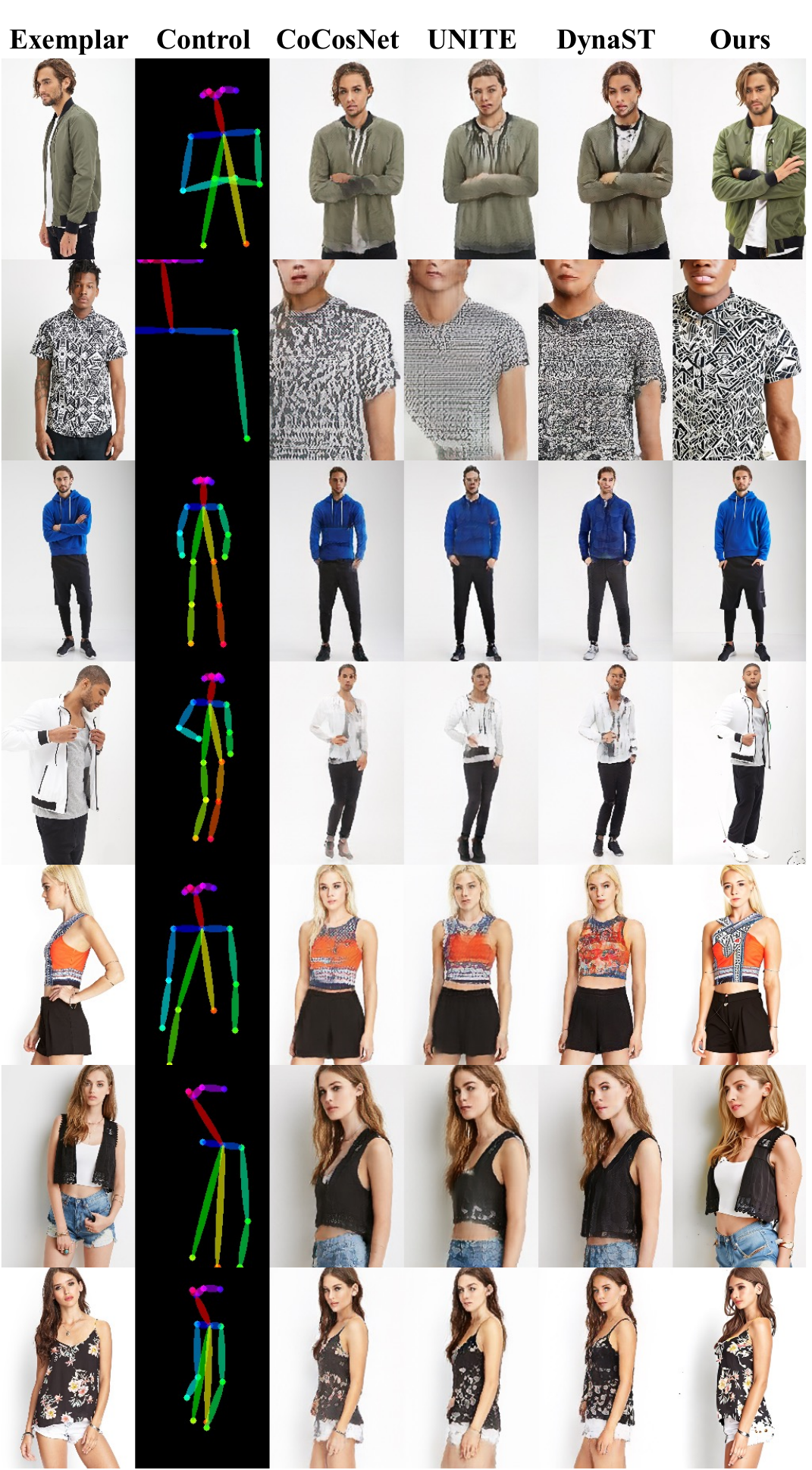}}
    \caption{\textbf{Pose-to-image} Qualitative comparisons on the DeepFashion Dataset.}
    \label{fig:sup3}
    \end{figure*}

%% file: supp/fig1-2.tex
\begin{figure*}[tb]
    \centering
    \makebox[\textwidth]{\includegraphics[width=\textwidth]{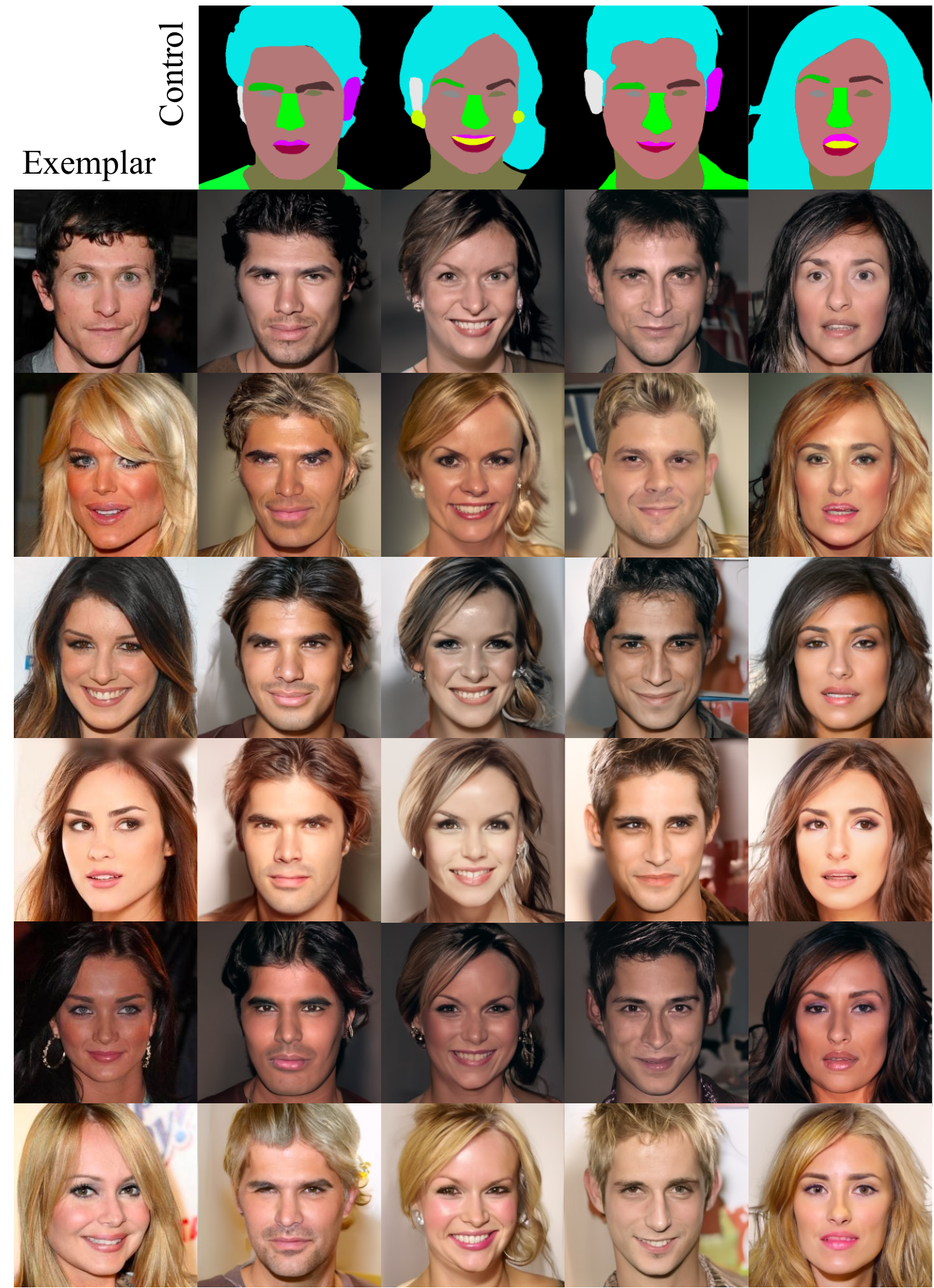}}
    \caption{\textbf{Mask-to-image} on the CelebAHQ Dataset.}
    \label{fig:sup1-2}
    \end{figure*}

%% file: supp/fig2-2.tex
\begin{figure*}[tb]
    \centering
    \makebox[\textwidth]{\includegraphics[width=\textwidth]{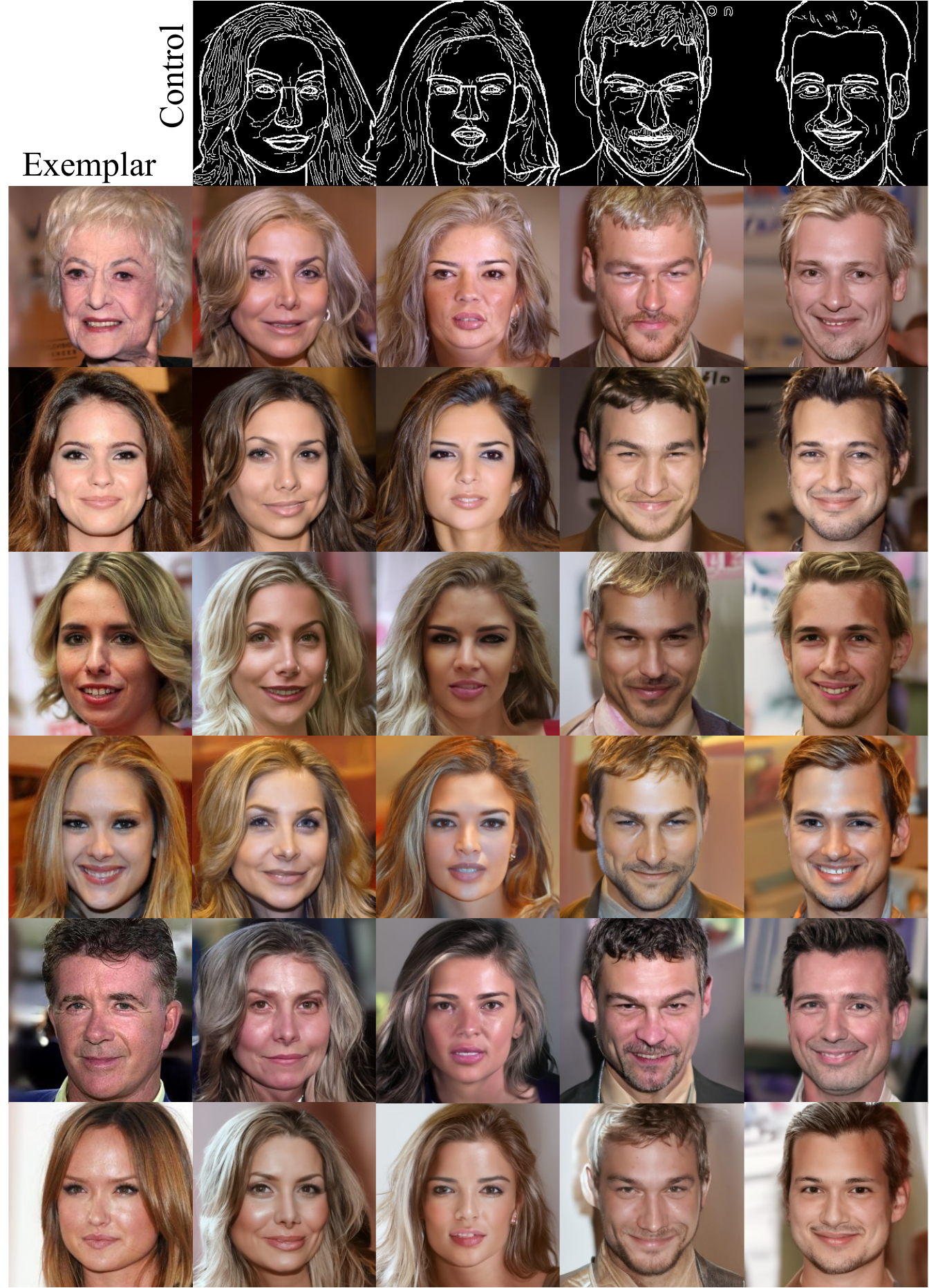}}
    \caption{\textbf{Edge-to-image} on the CelebA-HQ Dataset.}
    \label{fig:sup2-2}
    \end{figure*}

%% file: supp/fig3-2.tex
\begin{figure*}[tb]
    \centering
    \makebox[\textwidth]{\includegraphics[width=\textwidth]{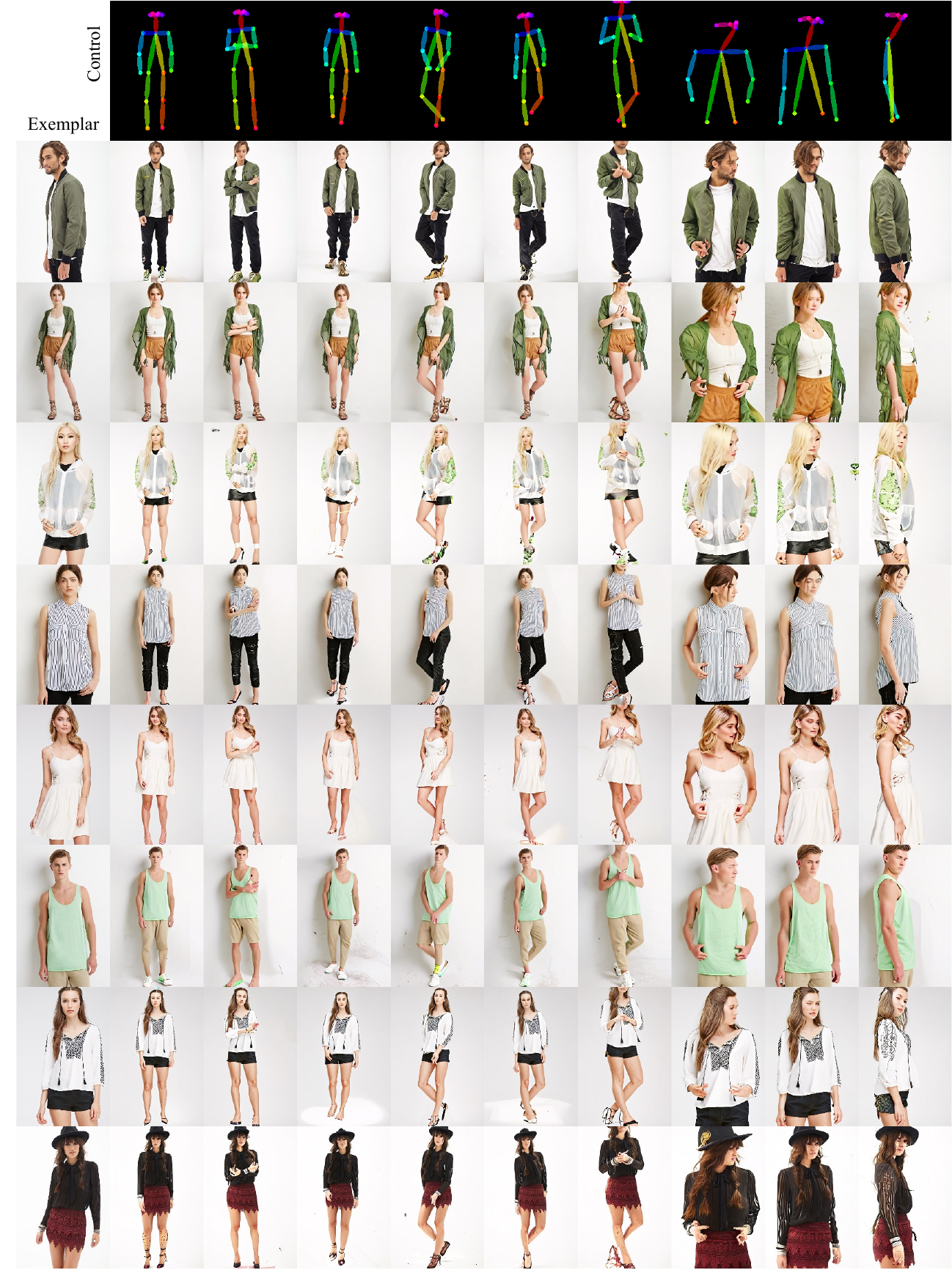}}
    \caption{\textbf{Pose-to-image} on the DeepFashion Dataset.}
    \label{fig:sup3-2}
    \end{figure*}